\documentclass{article}

\usepackage{microtype}
\usepackage{graphicx}
\usepackage{subcaption}
\usepackage{booktabs}
\usepackage{hyperref}

\usepackage[accepted]{icml2026}

\usepackage{amsmath}
\usepackage{amssymb}
\usepackage{mathtools}
\usepackage{amsthm}
\usepackage{stfloats}
\usepackage{placeins}

\usepackage{enumitem}

\usepackage{pifont}
\newcommand{\cmark}{\ding{51}}
\newcommand{\xmark}{\ding{55}}

\usepackage{multirow}

\usepackage{colortbl}
\definecolor{wingreen}{RGB}{220, 255, 220}  
\definecolor{lossred}{RGB}{255, 230, 230}   

\usepackage[capitalize,noabbrev]{cleveref}

\theoremstyle{plain}

\theoremstyle{definition}

\theoremstyle{remark}

\icmltitlerunning{Geometry-Aware Tabular Diffusion}

\begin{document}

\twocolumn[
\icmltitle{Geometry-Aware Tabular Diffusion}

\icmlsetsymbol{equal}{*}

\begin{icmlauthorlist}
\icmlauthor{David Turtora Zagardo}{gw}
\end{icmlauthorlist}

\icmlaffiliation{gw}{Independent Researcher}

\icmlcorrespondingauthor{David Turtora Zagardo}{dave@greenwillowstudios.com}

\icmlkeywords{Diffusion Models, Tabular Data, Geometric Deep Learning, Data Synthesis}

\vskip 0.3in
]

\printAffiliationsAndNotice{}

\begin{abstract}
Tabular synthesis is critical for privacy-preserving sharing and augmentation, yet diffusion models rely on implicit mechanisms to capture inter-column relationships. We introduce Geometry-Aware Tabular Diffusion (GATD), which augments tabular diffusion denoisers with pairwise angles and lengths computed from column value differences and used as inputs and auxiliary targets. Our MLP instantiation achieves state-of-the-art benchmark performance while using $3.5\times$ fewer parameters on average (up to $25\times$ for classification tasks): on ten datasets, it wins 8/10 Shape, 7/10 Trend, and 9/10 downstream utility (F1/RMSE), reducing Shape and Trend error by 27\% and 20\%. Default loss weights transfer to GNN and Transformer denoisers, improving Shape on 27/30 and Trend on 25/30 architecture-dataset cells. A matched ablation shows supervision (not extra inputs or capacity) drives the gain. This shows explicit relational supervision is a portable inductive bias for tabular diffusion.
\end{abstract}

\section{Introduction}
\label{sec:intro}

Tabular data remains the dominant format in enterprise applications, healthcare, and scientific research. The ability to synthesize realistic tabular data enables privacy-preserving data sharing~\citep{zhang2024tabsyn}, augmenting limited training sets~\citep{kotelnikov2023tabddpm}, and facilitating downstream model development without exposing sensitive records. However, tabular synthesis presents unique difficulties. Unlike images or text, tabular data exhibits heterogeneous column types, complex inter-column dependencies, and highly non-Gaussian marginal distributions---properties that have challenged deep generative models.

Diffusion models have recently emerged as a promising approach to tabular synthesis. Methods such as TabDDPM~\citep{kotelnikov2023tabddpm}, STaSy~\citep{kim2023stasy}, TabSyn~\citep{zhang2024tabsyn}, and TabDiff~\citep{shi2025tabdiff} adapt the denoising framework to mixed continuous-categorical data, achieving strong results on standard benchmarks. Among these, transformer-based architectures have become prevalent: self-attention mechanisms provide a flexible means to model relationships between columns, allowing the network to learn what should covary and how.

Yet this flexibility leaves inter-column structure to be inferred from the denoising objective alone. This raises a natural question: \emph{can we provide explicit relational structure as an auxiliary supervision signal, and does the same signal transfer across denoising architectures within tabular diffusion?}

\begin{figure*}[t]
\centering
\includegraphics[width=\textwidth]{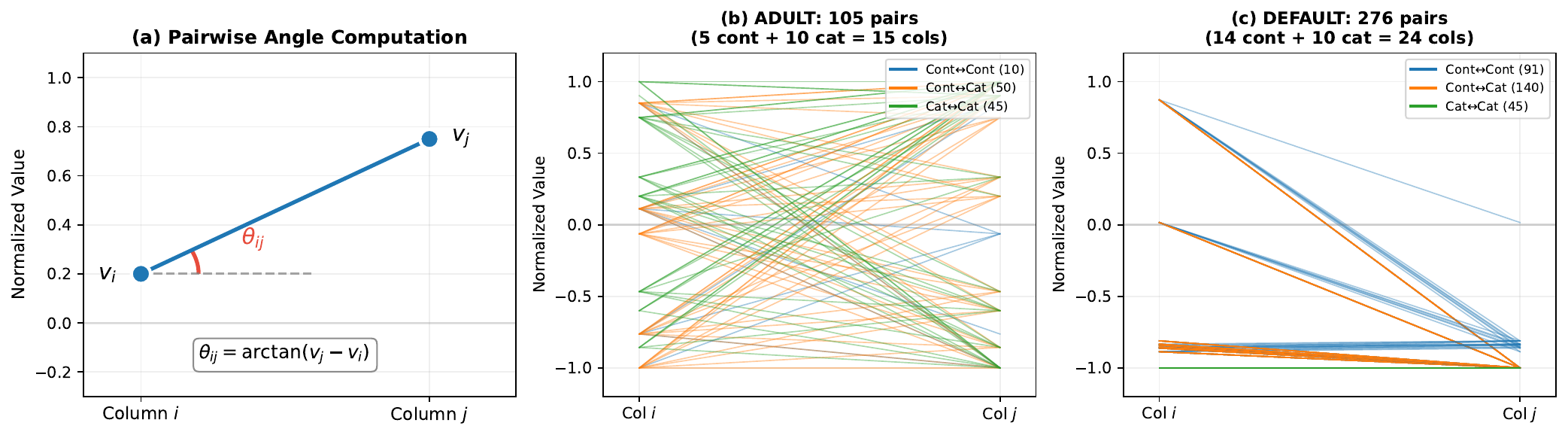}
\caption{\textbf{Geometric Intuition.} Inter-column relationships are encoded as pairwise angles $\theta_{ij} = \arctan(v_j - v_i)$ and lengths $\ell_{ij} = \frac{1}{2}\log(1 + (v_j - v_i)^2)$, providing explicit relational targets. Diagrams (b) and (c) show sample rows from \textbf{Adult} and \textbf{Default}.}
\label{fig:intuition}
\end{figure*}

We answer affirmatively to both. This paper introduces \textbf{Geometry-Aware Tabular Diffusion (GATD)}, which augments tabular diffusion denoisers with explicit pairwise geometric features computed directly from column values: an angle capturing the directional relationship between columns and a length capturing magnitude (Figure~\ref{fig:intuition}; full definitions in Section~\ref{sec:geometric_features}). These features are provided as model inputs, and crucially, the model is trained to predict them via auxiliary losses. The geometric representation provides an explicit encoding of inter-column structure that, we find, transfers across architecturally distinct diffusion denoisers.

Our claim is not that attention or message passing cannot learn such structure, but that explicit geometric supervision can reduce the burden on the denoiser and provide a portable relational inductive bias. 

A key finding is that geometric \emph{supervision} is essential, not merely geometric \emph{inputs}: an architecture-matched ablation shows that supplying geometric inputs and prediction heads without supervision yields no benefit (Cohen's $d = -0.08$), while restoring supervision produces a large effect ($d = 0.81$; Section~\ref{sec:capacity_control}). The auxiliary prediction task forces the network to internalize inter-column structure; architectural machinery alone produces no benefit.

We evaluate the geometric signal as a drop-in module across three diffusion denoising backbones: a residual Diffusion-MLP, a GNN with Laplacian-eigenmap positional encoding, and a column-wise Transformer. All use the same default geometric loss weights, $(\lambda_\theta,\lambda_\ell,\lambda_c)=(15,15,8)$, on ten benchmark datasets with 3 training seeds and 20 generation seeds per cell. Full per-architecture statistics and the MLP+Geom-vs-TabDiff comparison appear in Section~\ref{sec:main_results} and Section~\ref{sec:sota_comparison}. As a corollary of the cross-architecture portability claim, the compact MLP instantiation matches or exceeds TabDiff.

A previously-reported categorical-anchor mechanism on the MLP backbone ($\rho = 0.70$, $p = 0.025$) does not generalize across architectures (Section~\ref{sec:analysis}, Appendix~\ref{app:cat_analysis}): we characterize categorical structure as one operating regime among several, not a necessary condition for $+\textsc{Geom}$.

\paragraph{Contributions.} We contribute: (1) pairwise angle/length features for tabular diffusion, used as inputs and auxiliary targets; (2) an architecture-matched supervision ablation showing InputsOnly is indistinguishable from NoGeom ($d=-0.08$) while supervised geometry is large-effect ($d=0.81$); (3) portability across MLP, GNN, and Transformer denoisers (27/30 Shape, 25/30 Trend wins) with shared defaults; (4) an efficient MLP instantiation matching/exceeding TabDiff; and (5) practical guidance on $O(d^2)$ scaling, sampling, and loss weights.


\section{Related Work}
\label{sec:related}

\subsection{Diffusion Models for Tabular Data}

Diffusion models have emerged as a powerful alternative, offering stable training and strong distributional coverage.

\textbf{TabDDPM}~\citep{kotelnikov2023tabddpm} pioneered diffusion for tabular data, combining Gaussian diffusion for continuous columns with multinomial diffusion~\citep{hoogeboom2021multinomial} for categoricals. \textbf{STaSy}~\citep{kim2023stasy} used score-based methods with self-paced learning. \textbf{CoDi}~\citep{lee2023codi} proposed co-evolving contrastive diffusion with separate models for continuous and categorical columns. \textbf{TabSyn}~\citep{zhang2024tabsyn} introduced a VAE-then-diffusion approach, applying diffusion in a learned latent space.

\textbf{TabDiff}~\citep{shi2025tabdiff} unifies continuous and categorical diffusion by combining EDM~\citep{karras2022edm} for continuous columns with masked diffusion~\citep{austin2021d3pm} for categoricals, adding learnable per-column noise schedules and a transformer architecture for modeling column relationships. TabDiff achieves state-of-the-art results, outperforming prior methods (TabDDPM, STaSy, CoDi, TabSyn, CTGAN, TVAE) on 17 of 21 measures for 3 core metrics across 7 benchmark datasets; we therefore adopt TabDiff as our primary baseline.

We use the same diffusion losses (EDM for continuous, masked cross-entropy for categorical), holding the diffusion framework constant to isolate the contribution of explicit geometric supervision (Section~\ref{sec:method}) and the reflection-based boundary handling described below. Geometry as a drop-in signal also improves transformer-based denoisers on these benchmarks, indicating it complements rather than replaces attention.

\subsection{Geometric Deep Learning}

Geometric deep learning incorporates geometric structure into neural networks~\citep{bronstein2021geometric}. \textbf{GNNs} pass messages on graphs~\citep{kipf2017semi,velickovic2018gat}; \textbf{positional encodings} in transformers~\citep{vaswani2017attention,su2024roformer} demonstrate how geometric information guides attention.

A key insight is that explicit geometric structure accelerates learning and improves generalization. However, geometric deep learning has focused on inherently structured data---graphs, point clouds, molecules. Tabular data, despite meaningful column relationships, has not benefited from geometric approaches. We bridge this gap by constructing geometric features from the implicit relational structure in tabular data.

\subsection{Position in the Literature}

To our knowledge, no prior tabular generator provides explicit pairwise geometric supervision. CTGAN/TVAE~\citep{xu2019modeling}, TabDDPM~\citep{kotelnikov2023tabddpm}, and CoDi~\citep{lee2023codi} use MLP backbones with no explicit relational modeling; TabSyn~\citep{zhang2024tabsyn} and TabDiff~\citep{shi2025tabdiff} use transformer backbones that learn column relationships implicitly through attention. Our pairwise angle and length features as both inputs and auxiliary prediction targets enable strong performance across architecturally diverse backbones, including a compact MLP that matches or exceeds transformer-based SOTA. The same supervision signal also improves transformer-based denoisers (Section~\ref{sec:main_results}), indicating geometric supervision and attention are complementary rather than substitutable inductive biases for relational modeling.

\section{Method}
\label{sec:method}

\subsection{Preliminaries}
\label{sec:preliminaries}

\paragraph{Diffusion Framework.}
We adopt TabDiff's~\citep{shi2025tabdiff} diffusion framework exactly: EDM~\citep{karras2022edm} for continuous columns and masked diffusion~\citep{austin2021d3pm} for categoricals (with learnable per-column $k$). For continuous columns, the denoised output is:
\begin{equation}
D_\theta(\mathbf{x}; \sigma) = c_{\text{skip}}(\sigma)\, \mathbf{x} + c_{\text{out}}(\sigma)\, F_\theta\big(c_{\text{in}}(\sigma)\, \mathbf{x}; \sigma\big),
\end{equation}
where $F_\theta$ is the raw network and the preconditioning coefficients are $c_{\text{in}} = 1/\sqrt{\sigma^2 + \sigma_{\text{data}}^2}$, $c_{\text{skip}} = \sigma_{\text{data}}^2/(\sigma^2 + \sigma_{\text{data}}^2)$, and $c_{\text{out}} = \sigma \cdot \sigma_{\text{data}}/\sqrt{\sigma^2 + \sigma_{\text{data}}^2}$~\citep{karras2022edm}. We use learnable per-column $\rho$ for noise scheduling~\citep{shi2025tabdiff}. This deliberate choice isolates the contribution of our geometric features from diffusion modifications.

\paragraph{Notation.}
Consider a tabular dataset with $d_{\text{cont}}$ continuous columns and $d_{\text{cat}}$ categorical columns, with $d = d_{\text{cont}} + d_{\text{cat}}$ total columns. We denote normalized column values as $v \in [-1, 1]^d$.

\subsection{Geometric Feature Representation}
\label{sec:geometric_features}

The core contribution of our work is augmenting the diffusion model with explicit pairwise geometric features that capture inter-column relationships.

\subsubsection{Pairwise Angles}
\label{sec:angles}

For each pair of columns $(i, j)$ with $i < j$, we compute:
\begin{equation}
    \theta_{ij} = \arctan(v_j - v_i)
    \label{eq:angle}
\end{equation}
This angle captures the \emph{directional relationship} between columns. It is bounded ($\theta_{ij} \in (-\frac{\pi}{2}, \frac{\pi}{2})$) and antisymmetric ($\theta_{ji} = -\theta_{ij}$).

\subsubsection{Pairwise Lengths}
\label{sec:lengths}

We also compute the log-length for each pair:
\begin{equation}
    \ell_{ij} = \frac{1}{2} \log(1 + (v_j - v_i)^2)
    \label{eq:length}
\end{equation}
This captures the \emph{magnitude} of the difference between columns, with the logarithm compressing large differences.

An otherwise-identical raw-difference ablation gives similar but slightly weaker aggregate performance (Appendix~\ref{app:arctan_vs_diffs}), indicating that pairwise supervision is the primary mechanism while the bounded arctan parameterization provides more stable targets.

\subsubsection{Handling Mixed Types}
\label{sec:mixed_types}

To compute geometric features across both continuous and categorical columns, we map all columns to a unified normalized space $v \in [-1, 1]^d$:
\begin{align}
    v_{\text{cont}} &= 2 \cdot \text{QuantileTransform}(x_{\text{cont}}) - 1 \\
    v_{\text{cat}} &= 2 \cdot \frac{\text{index}}{\max(\text{cardinality} - 1, 1)} - 1
\end{align}
This enables geometric computation across all $\binom{d}{2}$ column pairs, regardless of type.

Representing categorical variables as continuous features has precedent in target encoding~\citep{miccibarreca2001preprocessing}, entity embeddings~\citep{guo2016entity}, and feature tokenization~\citep{gorishniy2021revisiting}. Our approach uses a fixed deterministic mapping to $[-1, 1]$ rather than learned embeddings, enabling direct computation of pairwise geometric features across all column types. A side benefit of this fixed mapping: ordinal columns (education levels, Likert scales) gain ordered reinforcement for free, since the deterministic position assignment preserves rank order, and small prediction errors yield neighboring categories rather than distant ones. For non-ordinal categoricals, the fixed ordering introduces a mild bias in the direction of error but does not affect predictive accuracy in our experiments.

\begin{figure*}[t]
\centering
\includegraphics[width=\textwidth]{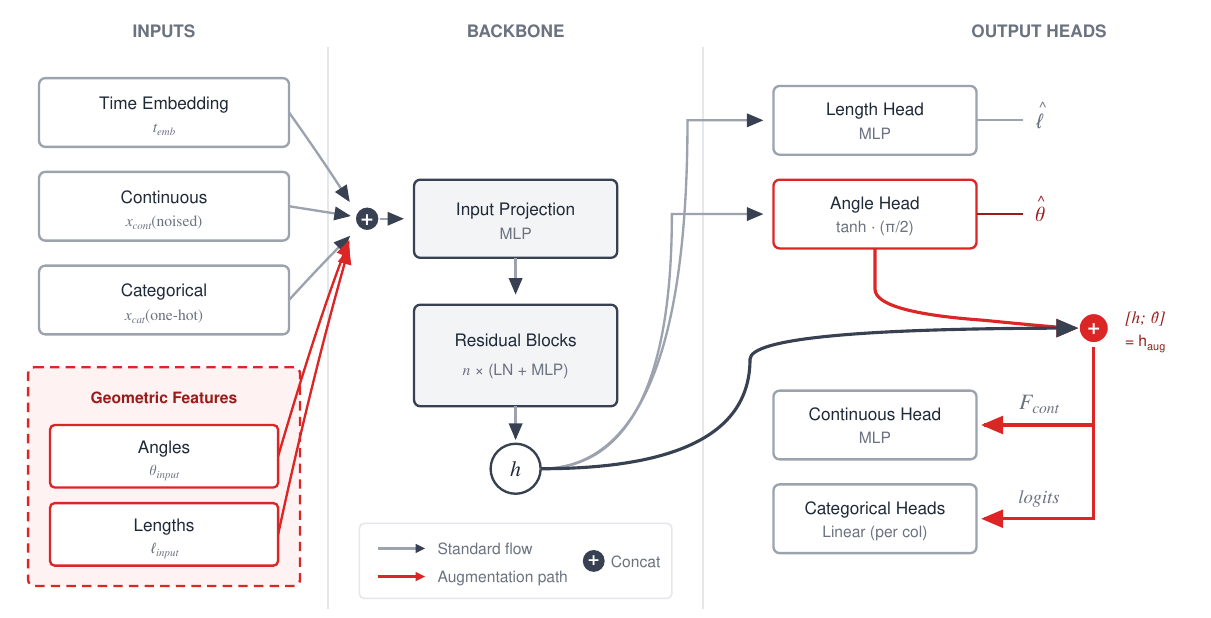}
\caption{\textbf{GeometryAwareMLP Architecture.} The model receives noised data and geometric features (angles $\boldsymbol{\theta}$, lengths $\boldsymbol{\ell}$) as input. After processing through residual blocks, geometric heads predict angles and lengths, which are concatenated with the hidden state before the denoising heads. This augmentation path (red) encourages the model to leverage geometric structure. It is critical to note that the length head is supervised during training, but detached during generation. The model only sees angles during sampling. The impact of this design choice is shown in Figure~\ref{fig:loss_ablation_violin}.}
\label{fig:architecture}
\end{figure*}

\subsection{Architecture}
\label{sec:architecture}

Our architecture, \textsc{GeometryAwareMLP}, extends a residual MLP with geometric inputs and auxiliary prediction heads (Figure~\ref{fig:architecture}).

\subsubsection{Input Representation}

The model receives a concatenated input:
\begin{equation}
    \mathbf{z}_{\text{input}} = [\mathbf{t}_{\text{emb}}; \mathbf{x}_{\text{cont}}; \mathbf{x}_{\text{cat}}^{\text{oh}}; \boldsymbol{\theta}_{\text{input}}; \boldsymbol{\ell}_{\text{input}}]
\end{equation}
where $\mathbf{t}_{\text{emb}} \in \mathbb{R}^{128}$ is a sinusoidal time embedding processed through a 2-layer MLP, $\mathbf{x}_{\text{cont}}$ contains the noised continuous values, $\mathbf{x}_{\text{cat}}^{\text{oh}}$ is the one-hot encoded categorical values, and $\boldsymbol{\theta}_{\text{input}}, \boldsymbol{\ell}_{\text{input}} \in \mathbb{R}^{d(d-1)/2}$ are geometric features computed from the current state.

\subsubsection{Network Backbone}

The input is processed through:
\begin{enumerate}[leftmargin=*,nosep]
    \item \textbf{Input projection:} 2-layer MLP mapping to hidden dimension $d_{\text{model}}$
    \item \textbf{Residual blocks:} $n_{\text{blocks}}$ residual MLP blocks with expansion factor 4 and dropout 0.1:
    \begin{equation}
        \text{ResidualBlock}(\mathbf{h}) = \mathbf{h} + \text{MLP}(\text{LayerNorm}(\mathbf{h}))
    \end{equation}
\end{enumerate}

\subsubsection{Output Heads}

\textbf{Geometric heads} predict from $\mathbf{h}$:
\begin{align}
    \hat{\boldsymbol{\theta}} &= \frac{\pi}{2} \cdot \tanh(\text{MLP}_\theta(\mathbf{h})) \\
    \hat{\boldsymbol{\ell}} &= \text{MLP}_\ell(\mathbf{h})
\end{align}
where each MLP includes LayerNorm, a hidden layer with GELU, and a linear output.

\textbf{Augmented representation:} Predicted angles are concatenated with the hidden state:
\begin{equation}
    \mathbf{h}_{\text{aug}} = [\mathbf{h}; \hat{\boldsymbol{\theta}}]
\end{equation}
We use angles only (not lengths) because angles encode strictly more information: 
$v_j - v_i = \tan(\theta_{ij})$ recovers both sign and magnitude, whereas lengths 
lose sign due to squaring. The length head provides auxiliary supervision to 
regularize the backbone, but its predictions are not used in $\mathbf{h}_{\text{aug}}$ 
to avoid redundant, potentially inconsistent signals.

\textbf{Denoising heads} predict from $\mathbf{h}_{\text{aug}}$:
\begin{align}
    F_{\text{cont}} &= \text{MLP}_{\text{cont}}(\mathbf{h}_{\text{aug}}) \\
    \text{logits}_c &= \text{Linear}(\text{LayerNorm}(\mathbf{h}_{\text{aug}}))
\end{align}

This augmentation encourages the model to leverage geometric structure when generating outputs.

\subsubsection{Architecture Comparison}

The MLP instantiation differs from TabDiff along four axes: column encoding (direct concatenation vs.\ learned embeddings), backbone architecture (residual MLP vs.\ transformer encoder--decoder), relationship modeling (explicit pairwise geometric features vs.\ self-attention), and parameter count ($\sim$400K--6M vs.\ $\sim$10M; $3.5\times$ fewer on average, up to $25\times$ for classification). The full component-level breakdown including shared diffusion machinery appears in Appendix~\ref{app:tabdiff_comparison}, Table~\ref{tab:tabdiff_detailed}.

\subsubsection{Cross-Architecture Variants}
\label{sec:cross_arch_variants}

The geometric-input, prediction-head, and augmentation-path mechanisms described above transfer to the GNN and Transformer diffusion denoisers evaluated in Section~\ref{sec:main_results}. The \textbf{GNN variant} replaces the residual MLP with edge-conditioned message passing on a complete column graph, with geometric features supplied as edge attributes, and pools node embeddings before the prediction heads. The \textbf{Transformer variant} uses a column-wise transformer encoder, with geometric features projected and concatenated to the denoiser representation. In all three diffusion backbones, the geometric prediction heads attach to the final pre-output representation and the augmentation path $\mathbf{h}_{\text{aug}} = [\mathbf{h}; \hat{\boldsymbol{\theta}}]$ feeds the denoising heads. Full architectural diagrams and per-backbone wiring details appear in Appendix~\ref{app:cross_arch_architectures}.

\subsubsection{Architecture Hyperparameters}

We use $d_{\text{model}} = 256$ throughout, with $n_{\text{blocks}} = 0$ for classification and $n_{\text{blocks}} = 8$ for regression (interpretation in Section~\ref{sec:analysis}).

\subsection{Training Objective}
\label{sec:loss}

Our loss combines standard diffusion objectives with geometric supervision:
\begin{equation}
    \mathcal{L} = \mathcal{L}_{\text{cont}} + \mathcal{L}_{\text{cat}} + \mathcal{L}_{\text{angle}} + \mathcal{L}_{\text{length}} + \mathcal{L}_{\text{consistency}}
\end{equation}

\subsubsection{Diffusion Losses}
\begin{equation}
    \mathcal{L}_{\text{diffusion}} = \lambda_\epsilon \mathcal{L}_{\text{cont}} + \lambda_{\text{cat}} \mathcal{L}_{\text{cat}}
\end{equation}
$\mathcal{L}_{\text{cont}}$ is EDM-weighted MSE~\citep{karras2022edm} on the denoised prediction. $\mathcal{L}_{\text{cat}}$ is cross-entropy for categorical columns~\citep{austin2021d3pm}, applied only to masked tokens and weighted by the inverse of the masking probability.

\subsubsection{Geometric Losses}
\begin{equation}
    \mathcal{L}_{\text{geometric}} = \lambda_\theta \mathcal{L}_{\text{angle}} + \lambda_\ell \mathcal{L}_{\text{length}} + \lambda_c \mathcal{L}_{\text{consistency}}
\end{equation}

\textbf{Angle/Length prediction:} Direct supervision on geometric predictions:
\begin{align}
    \mathcal{L}_{\text{angle}} &= \|\hat{\boldsymbol{\theta}} - \boldsymbol{\theta}_{\text{true}}\|^2 \\
    \mathcal{L}_{\text{length}} &= \|\hat{\boldsymbol{\ell}} - \boldsymbol{\ell}_{\text{true}}\|^2
\end{align}

\textbf{Consistency loss:} Ensures predicted geometry matches model outputs:
\begin{align}
    \mathcal{L}_{\text{consistency}} = \mathbb{E}[(1-t)^2] \cdot \big( 
    &\lVert\hat{\boldsymbol{\theta}} - \text{sg}(\boldsymbol{\theta}_{\text{pred}})\rVert^2 \nonumber \\
    &+ \lVert\hat{\boldsymbol{\ell}} - \text{sg}(\boldsymbol{\ell}_{\text{pred}})\rVert^2 \big)
\end{align}
where $\boldsymbol{\theta}_{\text{pred}}, \boldsymbol{\ell}_{\text{pred}}$ are computed from the denoised output and $\text{sg}(\cdot)$ denotes stop-gradient. The $(1-t)^2$ weighting emphasizes consistency at low noise levels, allowing the model to prioritize enforcing consistency when it is most valuable.

\subsubsection{Loss Weights}
We use $\lambda_\epsilon = 1.0$, $\lambda_{\text{cat}} = 0.05$, $\lambda_\theta = 15.0$, $\lambda_\ell = 15.0$, and $\lambda_c = 8.0$ for all datasets.

\paragraph{Loss Weight Analysis.} Our weighting inverts the typical loss hierarchy: at convergence, the weighted geometric terms ($\lambda_\theta \mathcal{L}_{\text{angle}} + \lambda_\ell \mathcal{L}_{\text{length}} + \lambda_c \mathcal{L}_{\text{consistency}}$) account for approximately 95\% of total loss, with diffusion losses contributing only 5\%. The inversion is essential: reducing geometric weights toward parity with diffusion degrades performance, suggesting heavy auxiliary supervision forces the network to internalize inter-column structure that transfers to improved generation. By contrast, TabDiff uses $\lambda_{\text{cat}} \approx 1.0$ (with optional annealing) and no geometric supervision.

\subsection{Training Details}
\label{sec:training}

We train with AdamW and EMA (hyperparameters in Table~\ref{tab:hyperparams}, Appendix~\ref{app:hyperparams}) for 20,000 epochs (vs.\ TabDiff's 8,000), with best model selected after epoch 10,000. GATD benefits from extended training while TabDiff does not (Appendix~\ref{app:tabdiff_epochs}). Despite $2.5\times$ more epochs, the complete training run is $1.7\times$ faster on average in wall-clock time; this is an end-to-end training-cost comparison. Sampling uses 1000 steps vs.\ TabDiff's 50 (Appendix~\ref{app:compute}).

\subsection{Sampling}
\label{sec:sampling}

At inference, we use EDM sampling (Euler method, 1000 steps) for continuous columns and iterative unmasking for categorical columns. Geometric features are computed during sampling but no geometric supervision is applied. The 1000-step setting is our high-fidelity operating point; reduced-step sampling preserves much of the gain (Appendix~\ref{app:steps_ablation}).

\paragraph{Boundary Handling.} During inverse transformation, generated continuous values are first mapped from the model's $[-1,1]$ scale to the quantile-transformer's $[0,1]$ scale. Let $s$ denote such a scaled value. Rather than hard clipping, we use reflection: if $s>1$ we map $s \mapsto 2-s$, and if $s<0$ we map $s \mapsto -s$, repeating up to 10 times before a final numerical clip to $[0,1]$. This avoids concentrating mass at the boundaries.

\begin{table}[hptb]   
\centering
\small
\begin{tabular}{llrrrrr}
\toprule
Dataset & Task & Train & Test & Cols & Cont & Cat \\
\midrule
Adult & C. & 32,561 & 16,281 & 15 & 5 & 10 \\
Beijing & R. & 37,581 & 4,176 & 12 & 7 & 5 \\
Bikesharing & R. & 15,641 & 1,738 & 13 & 5 & 8 \\
California & R. & 18,576 & 2,064 & 9 & 9 & 0 \\
Default & C. & 27,000 & 3,000 & 24 & 14 & 10 \\
Diabetes & C. & 61,059 & 20,354 & 37 & 9 & 28 \\
Magic & C. & 17,117 & 1,902 & 11 & 10 & 1 \\
News & R. & 35,679 & 3,965 & 48 & 43 & 5 \\
Powerplant & R. & 8,611 & 957 & 5 & 5 & 0 \\
Shoppers & C. & 11,097 & 1,233 & 18 & 7 & 11 \\
\bottomrule
\end{tabular}
\caption{Dataset statistics. We reclassify low-cardinality integers as categorical: education.num (Adult), Administrative, Informational, SpecialDay (Shoppers), and n\_tokens\_title, n\_non\_stop\_words, num\_keywords (News). We keep TabDiff's column classification the same for benchmarking reproducibility.}
\label{tab:datasets}
\end{table}


\section{Experiments}
\label{sec:experiments}

\subsection{Experimental Setup}
\label{sec:setup}

\paragraph{Datasets.}
We evaluate on seven datasets from the TabDiff benchmark and 3 additional datasets, spanning 5 binary classification and 5 regression tasks (Table~\ref{tab:datasets}).

\paragraph{Metrics.}
Following TabDiff, we evaluate: (1)~\textbf{Shape}---marginal distribution fidelity via SDMetrics~\citep{patki2016sdv} ColumnShapeSimilarity; (2)~\textbf{Trend}---correlation preservation via ColumnPairTrendsSimilarity; (3)~\textbf{MLE (machine learning efficacy)}---downstream utility using XGBoost~\citep{chen2016xgboost} (AUROC/F1 for classification, R$^2$/RMSE for regression).

\paragraph{Baselines.}
Our primary evaluation compares $+\textsc{Geom}$ against the same diffusion denoising architecture without geometric features or losses, across three backbones: a residual MLP (the original GATD), a GNN with Laplacian-eigenmap positional encoding (GNN+LE), and a column-wise Transformer. We additionally compare against \textbf{TabDiff}~\citep{shi2025tabdiff}, the prior state-of-the-art tabular diffusion model, on the MLP track to anchor absolute performance against published numbers (Section~\ref{sec:sota_comparison}). For the supervision-vs-capacity ablation, we further include an \textbf{InputsOnly} configuration defined in Section~\ref{sec:capacity_control}.

\paragraph{Protocol.}
We extend the TabDiff protocol: 20,000 training epochs (vs.\ TabDiff's 8,000), best model after epoch 10,000 (vs.\ 4,000), 3 train seeds, 20 generation seeds per train seed.

\subsection{Main Results: Cross-Architecture Evaluation}
\label{sec:main_results}

Two complementary pieces of evidence support our methodological claim. First, direct supervision is the operative variable: holding architecture, parameters, and gradient topology constant, removing only the geometric loss weights collapses performance to the no-geometry baseline (Section~\ref{sec:capacity_control}). Second, the same supervision signal ports across architecturally diverse diffusion backbones: Table~\ref{tab:fidelity_main} reports pairwise Shape and Trend error of $+\textsc{Geom}$ vs.\ its same-architecture baseline across three diffusion denoising backbones, applied as a drop-in module without architecture-specific tuning. Per-dataset MLE-1 and MLE-2 (downstream-utility) results appear in Tables~\ref{tab:mle1_appendix} and~\ref{tab:mle2_appendix} (Appendix~\ref{app:mle_per_dataset}); aggregate MLE behavior is summarized in the population-level test paragraph below.

\paragraph{Statistical results.} $+\textsc{Geom}$ wins 27/30 Shape and 25/30 Trend cells across the MLP, GNN, and Transformer diffusion backbones (per-architecture Shape 9/8/10 and Trend 8/9/8). Treating each architecture-dataset metric cell as a Bernoulli win/loss under a 50\% null win rate, the 52 wins out of 60 Shape/Trend cells give a two-sided exact sign-test value of $p = 5.21 \times 10^{-9}$. On downstream utility, $+\textsc{Geom}$ improves aggregate AUROC/R$^2$ on all three diffusion backbones; F1/RMSE improves on two of three. These results support the central claim that geometric supervision is portable across diffusion denoising architectures, while leaving non-diffusion generative frameworks to future work.

\begin{table*}[t]
\centering
\setlength{\tabcolsep}{4pt}
\renewcommand{\arraystretch}{1.1}
\caption{Cross-architecture fidelity: pairwise Shape and Trend error of $+\textsc{Geom}$ vs.\ its same-architecture baseline across three diffusion denoising backbones (MLP, GNN+LE, Transformer); 3 training seeds and 20 generation seeds per cell. \textbf{Bold} = better cell. $\Delta$ rows report relative improvement (\textcolor{blue}{$\uparrow$} better; \textcolor{red}{$\downarrow$} worse). $+\textsc{Geom}$ wins 27/30 Shape cells and 25/30 Trend cells.}
\label{tab:fidelity_main}
\resizebox{\textwidth}{!}{%
\begin{tabular}{lccccccccccc}
\toprule
\textbf{Method} & \textbf{Adult} & \textbf{Default} & \textbf{Diabetes} & \textbf{Magic} & \textbf{Shoppers} & \textbf{Beijing} & \textbf{Bikeshare} & \textbf{California} & \textbf{News} & \textbf{Power} & \textbf{Avg.} \\
\midrule
\multicolumn{12}{l}{\textit{Shape error $\downarrow$}} \\
MLP & 0.819\,{\scriptsize$\pm$0.0454} & 1.17\,{\scriptsize$\pm$0.0780} & 1.27\,{\scriptsize$\pm$0.0210} & 0.943\,{\scriptsize$\pm$0.107} & 1.58\,{\scriptsize$\pm$0.0821} & \textbf{0.797\,{\scriptsize$\pm$0.0424}} & 0.778\,{\scriptsize$\pm$0.0734} & 1.00\,{\scriptsize$\pm$0.0597} & 2.79\,{\scriptsize$\pm$0.0518} & 1.15\,{\scriptsize$\pm$0.173} & 1.23 \\
MLP+Geom & \textbf{0.493\,{\scriptsize$\pm$0.0442}} & \textbf{0.891\,{\scriptsize$\pm$0.0928}} & \textbf{0.472\,{\scriptsize$\pm$0.0208}} & \textbf{0.814\,{\scriptsize$\pm$0.0802}} & \textbf{0.845\,{\scriptsize$\pm$0.0725}} & 0.837\,{\scriptsize$\pm$0.0466} & \textbf{0.745\,{\scriptsize$\pm$0.0847}} & \textbf{0.963\,{\scriptsize$\pm$0.0451}} & \textbf{1.59\,{\scriptsize$\pm$0.0430}} & \textbf{0.973\,{\scriptsize$\pm$0.207}} & \textbf{0.862} \\
\hspace{1em}$\Delta$ & \textcolor{blue}{39.7\%}$\uparrow$ & \textcolor{blue}{23.6\%}$\uparrow$ & \textcolor{blue}{62.7\%}$\uparrow$ & \textcolor{blue}{13.6\%}$\uparrow$ & \textcolor{blue}{46.7\%}$\uparrow$ & \textcolor{red}{5.0\%}$\downarrow$ & \textcolor{blue}{4.2\%}$\uparrow$ & \textcolor{blue}{3.8\%}$\uparrow$ & \textcolor{blue}{43.0\%}$\uparrow$ & \textcolor{blue}{15.1\%}$\uparrow$ & \textcolor{blue}{29.8\%}$\uparrow$ \\
\cmidrule(lr){2-12}
GNN+LE & 0.940\,{\scriptsize$\pm$0.0637} & \textbf{1.40\,{\scriptsize$\pm$0.0821}} & 1.04\,{\scriptsize$\pm$0.0530} & \textbf{0.796\,{\scriptsize$\pm$0.133}} & 1.30\,{\scriptsize$\pm$0.0859} & 1.37\,{\scriptsize$\pm$0.103} & 0.989\,{\scriptsize$\pm$0.0875} & 0.885\,{\scriptsize$\pm$0.0462} & 2.54\,{\scriptsize$\pm$0.139} & 1.07\,{\scriptsize$\pm$0.196} & 1.23 \\
GNN+Geom & \textbf{0.614\,{\scriptsize$\pm$0.0731}} & 1.47\,{\scriptsize$\pm$0.124} & \textbf{0.826\,{\scriptsize$\pm$0.0601}} & 1.48\,{\scriptsize$\pm$0.275} & \textbf{1.13\,{\scriptsize$\pm$0.103}} & \textbf{0.858\,{\scriptsize$\pm$0.0699}} & \textbf{0.853\,{\scriptsize$\pm$0.0873}} & \textbf{0.872\,{\scriptsize$\pm$0.0464}} & \textbf{2.52\,{\scriptsize$\pm$0.338}} & \textbf{1.04\,{\scriptsize$\pm$0.193}} & \textbf{1.17} \\
\hspace{1em}$\Delta$ & \textcolor{blue}{34.7\%}$\uparrow$ & \textcolor{red}{5.0\%}$\downarrow$ & \textcolor{blue}{20.7\%}$\uparrow$ & \textcolor{red}{85.8\%}$\downarrow$ & \textcolor{blue}{13.1\%}$\uparrow$ & \textcolor{blue}{37.1\%}$\uparrow$ & \textcolor{blue}{13.8\%}$\uparrow$ & \textcolor{blue}{1.5\%}$\uparrow$ & \textcolor{blue}{0.8\%}$\uparrow$ & \textcolor{blue}{3.1\%}$\uparrow$ & \textcolor{blue}{5.4\%}$\uparrow$ \\
\cmidrule(lr){2-12}
\cmidrule(lr){2-12}
Transformer & 0.870\,{\scriptsize$\pm$0.0415} & 1.04\,{\scriptsize$\pm$0.0662} & 1.46\,{\scriptsize$\pm$0.0275} & 0.861\,{\scriptsize$\pm$0.0741} & 1.64\,{\scriptsize$\pm$0.101} & 1.06\,{\scriptsize$\pm$0.0418} & 0.876\,{\scriptsize$\pm$0.0813} & 0.892\,{\scriptsize$\pm$0.0603} & 2.41\,{\scriptsize$\pm$0.0409} & 1.17\,{\scriptsize$\pm$0.164} & 1.23 \\
Transformer+Geom & \textbf{0.832\,{\scriptsize$\pm$0.0478}} & \textbf{0.881\,{\scriptsize$\pm$0.0791}} & \textbf{0.697\,{\scriptsize$\pm$0.0232}} & \textbf{0.850\,{\scriptsize$\pm$0.0820}} & \textbf{1.26\,{\scriptsize$\pm$0.0765}} & \textbf{0.849\,{\scriptsize$\pm$0.0564}} & \textbf{0.807\,{\scriptsize$\pm$0.0770}} & \textbf{0.826\,{\scriptsize$\pm$0.0539}} & \textbf{1.63\,{\scriptsize$\pm$0.0633}} & \textbf{0.960\,{\scriptsize$\pm$0.198}} & \textbf{0.959} \\
\hspace{1em}$\Delta$ & \textcolor{blue}{4.3\%}$\uparrow$ & \textcolor{blue}{15.1\%}$\uparrow$ & \textcolor{blue}{52.3\%}$\uparrow$ & \textcolor{blue}{1.3\%}$\uparrow$ & \textcolor{blue}{23.0\%}$\uparrow$ & \textcolor{blue}{19.7\%}$\uparrow$ & \textcolor{blue}{7.9\%}$\uparrow$ & \textcolor{blue}{7.4\%}$\uparrow$ & \textcolor{blue}{32.3\%}$\uparrow$ & \textcolor{blue}{17.9\%}$\uparrow$ & \textcolor{blue}{21.8\%}$\uparrow$ \\
\midrule
\multicolumn{12}{l}{\textit{Trend error $\downarrow$}} \\
MLP & 1.75\,{\scriptsize$\pm$0.169} & 3.04\,{\scriptsize$\pm$0.774} & 2.48\,{\scriptsize$\pm$0.0593} & \textbf{1.08\,{\scriptsize$\pm$0.308}} & 2.23\,{\scriptsize$\pm$0.200} & \textbf{2.10\,{\scriptsize$\pm$0.229}} & 2.99\,{\scriptsize$\pm$0.397} & 1.34\,{\scriptsize$\pm$0.172} & 1.57\,{\scriptsize$\pm$0.210} & 0.886\,{\scriptsize$\pm$0.212} & 1.95 \\
MLP+Geom & \textbf{1.23\,{\scriptsize$\pm$0.116}} & \textbf{2.93\,{\scriptsize$\pm$0.936}} & \textbf{1.43\,{\scriptsize$\pm$0.0562}} & 1.10\,{\scriptsize$\pm$0.285} & \textbf{1.83\,{\scriptsize$\pm$0.190}} & 2.17\,{\scriptsize$\pm$0.221} & \textbf{2.86\,{\scriptsize$\pm$0.401}} & \textbf{1.33\,{\scriptsize$\pm$0.403}} & \textbf{1.20\,{\scriptsize$\pm$0.222}} & \textbf{0.387\,{\scriptsize$\pm$0.147}} & \textbf{1.65} \\
\hspace{1em}$\Delta$ & \textcolor{blue}{29.6\%}$\uparrow$ & \textcolor{blue}{3.8\%}$\uparrow$ & \textcolor{blue}{42.2\%}$\uparrow$ & \textcolor{red}{2.1\%}$\downarrow$ & \textcolor{blue}{18.1\%}$\uparrow$ & \textcolor{red}{3.2\%}$\downarrow$ & \textcolor{blue}{4.2\%}$\uparrow$ & \textcolor{blue}{0.3\%}$\uparrow$ & \textcolor{blue}{23.3\%}$\uparrow$ & \textcolor{blue}{56.4\%}$\uparrow$ & \textcolor{blue}{15.4\%}$\uparrow$ \\
\cmidrule(lr){2-12}
GNN+LE & 0.994\,{\scriptsize$\pm$0.0843} & 1.97\,{\scriptsize$\pm$0.165} & 3.30\,{\scriptsize$\pm$0.434} & 20.2\,{\scriptsize$\pm$13.4} & 2.41\,{\scriptsize$\pm$0.336} & 2.76\,{\scriptsize$\pm$0.440} & \textbf{3.35\,{\scriptsize$\pm$0.612}} & 8.30\,{\scriptsize$\pm$0.824} & 13.5\,{\scriptsize$\pm$0.205} & 0.372\,{\scriptsize$\pm$0.116} & 5.72 \\
GNN+Geom & \textbf{0.984\,{\scriptsize$\pm$0.0819}} & \textbf{1.80\,{\scriptsize$\pm$0.515}} & \textbf{2.94\,{\scriptsize$\pm$0.180}} & \textbf{8.72\,{\scriptsize$\pm$4.74}} & \textbf{1.96\,{\scriptsize$\pm$0.218}} & \textbf{2.53\,{\scriptsize$\pm$0.462}} & 3.42\,{\scriptsize$\pm$0.459} & \textbf{7.23\,{\scriptsize$\pm$0.996}} & \textbf{12.0\,{\scriptsize$\pm$0.519}} & \textbf{0.263\,{\scriptsize$\pm$0.120}} & \textbf{4.19} \\
\hspace{1em}$\Delta$ & \textcolor{blue}{0.9\%}$\uparrow$ & \textcolor{blue}{8.7\%}$\uparrow$ & \textcolor{blue}{11.0\%}$\uparrow$ & \textcolor{blue}{56.8\%}$\uparrow$ & \textcolor{blue}{18.5\%}$\uparrow$ & \textcolor{blue}{8.4\%}$\uparrow$ & \textcolor{red}{2.1\%}$\downarrow$ & \textcolor{blue}{12.9\%}$\uparrow$ & \textcolor{blue}{11.0\%}$\uparrow$ & \textcolor{blue}{29.2\%}$\uparrow$ & \textcolor{blue}{26.7\%}$\uparrow$ \\
\cmidrule(lr){2-12}
\cmidrule(lr){2-12}
Transformer & 1.70\,{\scriptsize$\pm$0.166} & 3.27\,{\scriptsize$\pm$1.05} & 2.91\,{\scriptsize$\pm$0.0512} & \textbf{1.12\,{\scriptsize$\pm$0.265}} & \textbf{2.15\,{\scriptsize$\pm$0.237}} & 2.30\,{\scriptsize$\pm$0.204} & 3.04\,{\scriptsize$\pm$0.430} & 1.35\,{\scriptsize$\pm$0.104} & 1.63\,{\scriptsize$\pm$0.234} & 0.978\,{\scriptsize$\pm$0.212} & 2.04 \\
Transformer+Geom & \textbf{1.65\,{\scriptsize$\pm$0.210}} & \textbf{2.90\,{\scriptsize$\pm$0.727}} & \textbf{1.81\,{\scriptsize$\pm$0.0977}} & 1.17\,{\scriptsize$\pm$0.265} & 2.21\,{\scriptsize$\pm$0.195} & \textbf{2.24\,{\scriptsize$\pm$0.224}} & \textbf{3.02\,{\scriptsize$\pm$0.406}} & \textbf{1.29\,{\scriptsize$\pm$0.283}} & \textbf{1.44\,{\scriptsize$\pm$0.242}} & \textbf{0.352\,{\scriptsize$\pm$0.141}} & \textbf{1.81} \\
\hspace{1em}$\Delta$ & \textcolor{blue}{2.9\%}$\uparrow$ & \textcolor{blue}{11.3\%}$\uparrow$ & \textcolor{blue}{37.7\%}$\uparrow$ & \textcolor{red}{4.3\%}$\downarrow$ & \textcolor{red}{3.0\%}$\downarrow$ & \textcolor{blue}{2.5\%}$\uparrow$ & \textcolor{blue}{0.7\%}$\uparrow$ & \textcolor{blue}{4.7\%}$\uparrow$ & \textcolor{blue}{11.4\%}$\uparrow$ & \textcolor{blue}{64.0\%}$\uparrow$ & \textcolor{blue}{11.5\%}$\uparrow$ \\
\bottomrule
\end{tabular}%
}
\end{table*}

\subsection{Architecture-Matched Capacity Control: Inputs vs Supervision}
\label{sec:capacity_control}

\begin{figure*}[t]
\centering
\includegraphics[width=\textwidth]{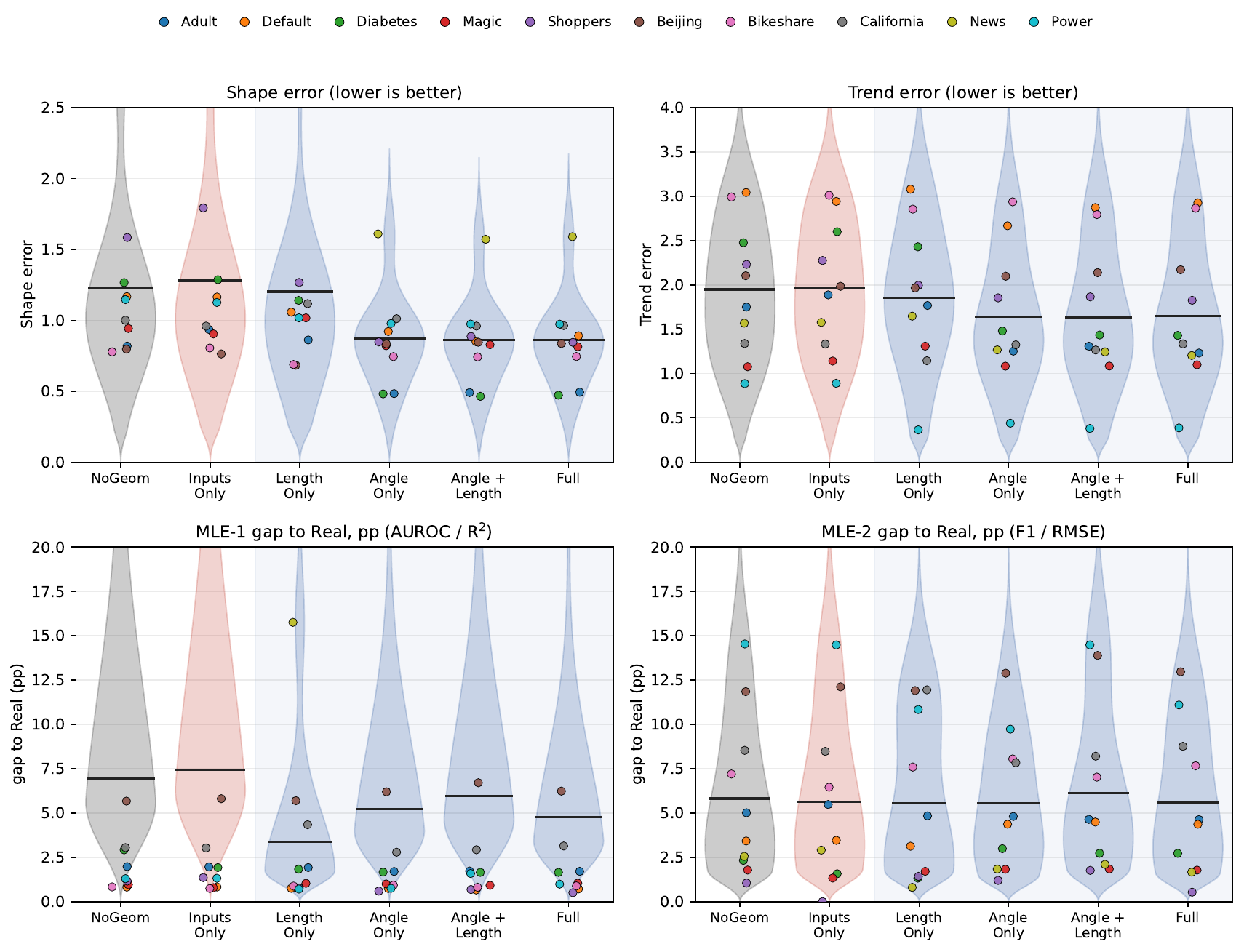}
\caption{\textbf{Loss-ablation distributions by method.} Each panel plots 10 dataset values across 6 methods using KDE violins, median bars, and colored dataset dots tracked across positions. The shaded band marks geometric supervision: medians drop sharply from Inputs Only to supervised methods for Shape (1.07 $\to$ 0.85) and Trend (1.93 $\to$ 1.40), while similar supervised violins show Angle Only / Angle + Length / Full have small aggregate fidelity differences. MLE-1 isolates News (yellow) as the main consistency-loss beneficiary, falling from $\sim$50--57pp to 30--40pp; MLE-2 remains similar, matching saturated downstream utility. Length-only provides the best News utility at the cost of fidelity, pulling the challenging dataset's MLE-1 score into the visible range $\sim$15--20pp.}
\label{fig:loss_ablation_violin}
\end{figure*}

The cross-architecture results could reflect added inputs/heads rather than supervision, so we compare three MLP configurations: \textbf{NoGeom} (no geometric inputs, heads, or losses), \textbf{InputsOnly} (full $+\textsc{Geom}$ architecture and augmentation path, but $\lambda_\theta=\lambda_\ell=\lambda_c=0$), and \textbf{Full $+\textsc{Geom}$}. InputsOnly and Full share parameter count, forward compute, and gradient topology; only direct geometric supervision differs. InputsOnly achieves lower total loss, yet its geometric losses remain 8--339$\times$ higher (Figure~\ref{fig:training_curves}); aggregate Shape errors are 1.229/1.279/0.862 for NoGeom/InputsOnly/Full. Pooled effects are $d(\textsc{NoGeom}\to\textsc{InputsOnly})=-0.08$, $d(\textsc{NoGeom}\to\textsc{Full})=+0.79$, and $d(\textsc{InputsOnly}\to\textsc{Full})=+0.81$. Thus supervision, not added capacity or augmentation path, drives the gain; Figure~\ref{fig:loss_ablation_violin} shows the corresponding cliff.

\begin{table}[t]
\centering
\caption{MLP+Geom vs. TabDiff summary. Errors and gaps are averaged over 10 datasets; lower is better. Full per-dataset values appear in Appendix~\ref{app:sota_per_dataset}.}
\label{tab:sota_summary}
\small
\begin{tabular}{lccc}
\toprule
Metric & TabDiff & MLP+Geom & Summary \\
\midrule
Shape error & 1.19 & \textbf{0.862} & 27.3\% lower \\
Trend error & 2.05 & \textbf{1.65} & 19.6\% lower \\
MLE-1 gap & 11.4\% & \textbf{4.8\%} & 58.2\% gap closure \\
MLE-2 gap & 9.7\% & \textbf{5.6\%} & 42.0\% gap closure \\
\bottomrule
\end{tabular}
\end{table}

\subsection{State-of-the-Art Comparison: MLP+Geom vs TabDiff}
\label{sec:sota_comparison}

As a corollary of the cross-architecture portability claim, we anchor the MLP instantiation of GATD against the prior state-of-the-art tabular diffusion model, TabDiff~\citep{shi2025tabdiff}. TabDiff results are reproduced from the official TabDiff codebase under their published 8,000-epoch protocol with three training seeds and twenty generation seeds per dataset. Two protocol differences could in principle confound the comparison; both are controlled in the appendix. Training-epoch budget: a matched 8,000-epoch run of GATD-MLP retains the gain (27/40 wins, 0.854 Shape vs.\ TabDiff-8k's 1.187 --- 28\% reduction, vs.\ 27\% at 20k; Appendix~\ref{app:gatd_8k_epochs}). Sampling-step budget: at 50 steps GATD already beats TabDiff on 3 of 4 aggregate metrics, and wins 4 of 4 at 100 steps (Figure~\ref{fig:step_ablation_summary}, Appendix~\ref{app:steps_ablation}).

\paragraph{Per-dataset wins.} MLP+Geom wins 8/10 on Shape, 7/10 on Trend, 6/10 on AUROC/R$^2$ (MLE-1), 9/10 on F1/RMSE (MLE-2). Only MLE-2 reaches 10-cell sign-test significance ($p = 0.022$); Shape ($p = 0.11$), Trend ($p = 0.34$), MLE-1 ($p = 0.75$) are directionally favorable but underpowered at $n=10$.

\paragraph{Aggregate gap-to-real closure.} Using train-on-real, test-on-real (TRTR) performance as the upper bound (Real rows of Tables~\ref{tab:mle1_appendix} and \ref{tab:mle2_appendix}), MLP+Geom closes 58.2\% of TabDiff's aggregate AUROC/R$^2$ gap to real ($11.4\% \to 4.8\%$) and 42.0\% of its F1/RMSE gap ($9.7\% \to 5.6\%$); per-dataset Shape and Trend errors are reduced by 27.3\% and 19.6\% on average. Largest wins (63.1pp on News for AUROC/R$^2$, 15.0pp on Beijing for F1/RMSE) substantially exceed largest losses (0.2pp on Adult, 0.5pp on California).

\begin{table}[t]
\centering
\small
\caption{Protocol-control summary for MLP+Geom vs.\ TabDiff. Lower is better. MLE-1 and MLE-2 report gap to real-data performance. The 8k row matches TabDiff's training budget; the 50/100-step rows test the sampling-budget concern. Step rows use the Appendix~\ref{app:steps_ablation} protocol.}
\label{tab:protocol_controls_main}
\setlength{\tabcolsep}{3.2pt}
\begin{tabular}{lcccc}
\toprule
Setting & Shape & Trend & MLE-1 & MLE-2 \\
\midrule
TabDiff default & 1.187 & 2.048 & 11.4\% & 9.7\% \\
GATD, 8k train & \textbf{0.854} & \textbf{1.680} & \textbf{4.7\%} & \textbf{6.1\%} \\
GATD, 50 steps & 1.546 & \textbf{1.957} & \textbf{4.6\%} & \textbf{7.2\%} \\
GATD, 100 steps & \textbf{1.125} & \textbf{1.714} & \textbf{4.9\%} & \textbf{5.4\%} \\
GATD default & \textbf{0.862} & \textbf{1.647} & \textbf{4.8\%} & \textbf{5.6\%} \\
\bottomrule
\end{tabular}
\end{table}
\paragraph{Calibrating the magnitude of improvement.} TabDiff's reported gain over the prior SOTA (TabSyn) was 13\% Shape and 23\% Trend reduction on 7 datasets, single seed~\citep{shi2025tabdiff}; GATD-MLP's 27.3\%/19.6\% reductions over TabDiff are on 10 datasets with $3 \times 20$ seeds. The cross-architecture evidence adds two additional diffusion backbones reaching SOTA-competitive Shape/Trend, neither previously demonstrated here as a drop-in tabular diffusion denoiser. The two cells where MLP+Geom under-performs TabDiff differ by very different magnitudes: the Magic Shape gap (0.4 percentage points) is well within one standard deviation (Table~\ref{tab:fidelity_main}, $\sigma \approx 0.08$) and is a tie within noise; the California Shape gap (22.8 percentage points) is a real loss. The 3-seed protocol additionally surfaces substantial training-seed instability in TabDiff that single-seed reporting would obscure (Table~\ref{tab:tabdiff_stability}); each $+\textsc{Geom}$ backbone is more reproducible than TabDiff on the majority of datasets. 

\subsection{Analysis}
\label{sec:analysis}

\paragraph{Architecture.}
Classification in MLP architectures favors $n_{\text{blocks}}=0$ (10/10 fidelity wins) while regression favors $n_{\text{blocks}}=8$ (10/10 MLE wins), both at $d_{\text{model}}=256$ (Appendix~\ref{app:architecture}). California and Powerplant (both 0 categorical columns) prefer 0 blocks even as regression tasks.

\paragraph{When Does Geometry Help Most?} On the MLP backbone, per-dataset Shape improvement is rank-correlated with categorical-column count ($\rho = 0.70$, $p = 0.025$); this is the empirical basis for the ``categorical anchor'' interpretation in the original GATD analysis. The same correlation is weaker and individually non-significant for the GNN and Transformer backbones ($\rho \in [0.38, 0.42]$, $p \geq 0.22$); pooled across the three diffusion backbones it reaches $\rho = 0.51$, $p = 0.004$. Continuous-only datasets such as Powerplant ($d=5$, 0 cat.\ cols.) nevertheless show gains under geometric supervision, demonstrating that categorical structure is one operating regime rather than a necessary condition. Per-architecture correlations and the categorical-fraction control analysis are reported in Appendix~\ref{app:cat_analysis}.

\paragraph{Computational Profile.}
Geometric features add $O(d^2)$ tensor operations per row. In our benchmark regime ($d \leq 48$), this cost is dominated by backbone forward/backward passes, and the all-pair structure is lesser or comparable in order to column-wise self-attention. Combined with shallow networks for classification ($n_{\text{blocks}}=0$), GATD-MLP's end-to-end training run is \textbf{1.7$\times$ faster} on average than TabDiff despite $2.5\times$ more epochs; this is a total wall-clock training comparison, not an epoch/second throughput claim (Appendix~\ref{app:compute}). The pairwise computation is already batched GPU-accelerated tensor operations. On APS ($d=171$), the profile becomes more memory-bound, but GATD remains faster per epoch and both models run on a single T4 (Appendix~\ref{app:aps_scalability}).

\begin{table}[hptb]
\centering
\small
\resizebox{\columnwidth}{!}{%
\begin{tabular}{lcccc}
\toprule
Method & Shape & Trend & MLE-1 & MLE-2 \\
\midrule
MLP+Geom & 1.84$\times$ (9/10) & 0.95$\times$ (5/10) & 2.52$\times$ (9/10) & 1.86$\times$ (8/10) \\
GNN+Geom & 0.66$\times$ (3/10) & 0.55$\times$ (2/10) & 1.28$\times$ (6/10) & 1.49$\times$ (6/10) \\
Transformer+Geom & 1.65$\times$ (8/10) & 1.37$\times$ (7/10) & 1.29$\times$ (5/10) & 1.29$\times$ (6/10) \\
\bottomrule
\end{tabular}%
}
\caption{Training-seed stability of $+\textsc{Geom}$ backbones relative to TabDiff. Each cell reports the median across-train-seed CV ratio (TabDiff~/~method) over 10 datasets, with the count of datasets where the method's std is smaller than TabDiff's in parentheses. CV ratios $>1$ and high parenthetical counts indicate the method is more stable than TabDiff.}
\label{tab:tabdiff_stability}
\end{table}

\section{Conclusion}
\label{sec:conclusion}

We introduced Geometry-Aware Tabular Diffusion (GATD), a portable relational inductive bias for tabular diffusion: explicit pairwise geometric features as both inputs and auxiliary prediction targets, evaluated as a drop-in module across MLP, GNN, and Transformer diffusion denoisers. Population-level statistical evidence appears in Section~\ref{sec:main_results}.

The architecture-matched ablation localizes the gain to direct supervision of the geometric heads (Section~\ref{sec:capacity_control}); the MLP instantiation matches or exceeds TabDiff without an attention mechanism (Section~\ref{sec:sota_comparison}). Cross-architecture results (Section~\ref{sec:main_results}) show that the same signal also improves transformer- and message-passing-based diffusion denoisers, indicating geometric supervision, attention, and message passing are complementary rather than mutually exclusive inductive biases.

\paragraph{Limitations.}
\textbf{$O(d^2)$ pairwise scaling.} The geometric signal is quadratic in column count because it intentionally supervises all column pairs, including weak or uncorrelated pairs that provide negative relational evidence. This is a practical memory consideration rather than a methodological barrier: column-wise self-attention is also $O(d^2)$ in the number of columns, and in our evaluated regime ($d \leq 48$) the cost is dominated by backbone compute. Wider-table behavior, characterized on APS Failure at Scania Trucks ($d = 171$; Appendix~\ref{app:aps_scalability}), shows the tradeoff directly: GATD's memory and parameter count exceed TabDiff's at this scale, but both models fit on a single T4 and minimal loss-weight tuning recovers 3 of 4 wins.

\textbf{Smaller downstream-utility effects.} F1/RMSE shows no reliable population-level cross-architecture signal; AUROC/R$^2$ reaches significance only under the magnitude-weighted Wilcoxon test. Some continuous-heavy regression cells with GNN or Transformer backbones regress in downstream utility, indicating that geometric supervision may require tuning when paired with relational backbones on continuous-heavy data.

\textbf{Empirical portability scope.} We evaluate three denoising backbones within the diffusion framework. Because the geometric features and losses are computed from data rather than from the diffusion mechanism itself, extending them to non-diffusion paradigms such as autoregressive models, VAEs, or normalizing flows is a plausible direction, but it is not directly supported by the experiments in this paper.

\textbf{Heuristic loss weights and operating-point choices.} The defaults $\lambda_\theta = \lambda_\ell = 15$, $\lambda_c = 8$ are heuristic; per-dataset tuning unlocks substantial gains on specific datasets (Section~\ref{sec:practitioner_tuning}, Figure~\ref{fig:lambda_sweep}, Appendix~\ref{app:lambda_sensitivity}). The 1000-step sampling budget is an operating point, not a requirement (Appendix~\ref{app:steps_ablation}).

\paragraph{Future Work.}
Promising directions include: (1) higher-order geometric supervision, such as triangle-closure losses on column triples; (2) learnable geometric transformations beyond fixed arctan/log-length; (3) extension to conditional generation and non-diffusion generative paradigms, evaluated separately from the present diffusion study; (4) integration with differentially-private training, where auxiliary supervision may interact with gradient clipping and noise injection; (5) Bayesian formulations for learning angle and length priors, including radial/directional posterior decompositions over magnitude and orientation~\citep{oh2020radial}; and (6) per-architecture loss-weight tuning.

\section*{Impact Statement}
This paper presents work on synthetic tabular data generation. While synthetic data is often motivated by privacy concerns, our method, like most generative models, provides no formal privacy guarantees. Organizations should not treat synthetic data as inherently privacy-preserving or as a substitute for rigorous techniques such as differential privacy. There is a risk that synthetic data generation could be used for ``privacy-washing''---presenting a privacy-conscious image to stakeholders without providing meaningful protection. We encourage practitioners to pair synthetic data generation methods with formal, mathematical privacy frameworks and training protocols like DPSGD~\citep{abadi2016deep} when handling sensitive information.

\section*{Acknowledgements}
The author thanks the ICML reviewers and area chair for constructive feedback that improved the experimental controls, scalability discussion, and presentation.

\bibliography{references}

\begin{thebibliography}{21}
\providecommand{\natexlab}[1]{#1}
\providecommand{\url}[1]{\texttt{#1}}
\expandafter\ifx\csname urlstyle\endcsname\relax
  \providecommand{\doi}[1]{doi: #1}\else
  \providecommand{\doi}{doi: \begingroup \urlstyle{rm}\Url}\fi

\bibitem[Abadi et~al.(2016)Abadi, Chu, Goodfellow, McMahan, Mironov, Talwar,
  and Zhang]{abadi2016deep}
Abadi, M., Chu, A., Goodfellow, I., McMahan, H.~B., Mironov, I., Talwar, K.,
  and Zhang, L.
\newblock Deep learning with differential privacy.
\newblock In \emph{Proceedings of the 2016 ACM SIGSAC Conference on Computer
  and Communications Security}, pp.\  308--318, 2016.

\bibitem[Austin et~al.(2021)Austin, Johnson, Ho, Tarlow, and van~den
  Berg]{austin2021d3pm}
Austin, J., Johnson, D.~D., Ho, J., Tarlow, D., and van~den Berg, R.
\newblock Structured denoising diffusion models in discrete state-spaces.
\newblock In \emph{Advances in Neural Information Processing Systems},
  volume~34, pp.\  17981--17993, 2021.

\bibitem[Bronstein et~al.(2021)Bronstein, Bruna, Cohen, and
  Veli{\v{c}}kovi{\'c}]{bronstein2021geometric}
Bronstein, M.~M., Bruna, J., Cohen, T., and Veli{\v{c}}kovi{\'c}, P.
\newblock Geometric deep learning: Grids, groups, graphs, geodesics, and
  gauges.
\newblock \emph{arXiv preprint arXiv:2104.13478}, 2021.

\bibitem[Chen \& Guestrin(2016)Chen and Guestrin]{chen2016xgboost}
Chen, T. and Guestrin, C.
\newblock {XGBoost}: A scalable tree boosting system.
\newblock In \emph{Proceedings of the 22nd ACM SIGKDD International Conference
  on Knowledge Discovery and Data Mining}, pp.\  785--794. ACM, 2016.
\newblock \doi{10.1145/2939672.2939785}.

\bibitem[Gorishniy et~al.(2021)Gorishniy, Rubachev, Khrulkov, and
  Babenko]{gorishniy2021revisiting}
Gorishniy, Y., Rubachev, I., Khrulkov, V., and Babenko, A.
\newblock Revisiting deep learning models for tabular data.
\newblock In \emph{Advances in Neural Information Processing Systems},
  volume~34, pp.\  18932--18943, 2021.

\bibitem[Guo \& Berkhahn(2016)Guo and Berkhahn]{guo2016entity}
Guo, C. and Berkhahn, F.
\newblock Entity embeddings of categorical variables.
\newblock \emph{arXiv preprint arXiv:1604.06737}, 2016.

\bibitem[Hoogeboom et~al.(2021)Hoogeboom, Nielsen, Jaini, Forr{\'e}, and
  Welling]{hoogeboom2021multinomial}
Hoogeboom, E., Nielsen, D., Jaini, P., Forr{\'e}, P., and Welling, M.
\newblock Argmax flows and multinomial diffusion: Learning categorical
  distributions.
\newblock In \emph{Advances in Neural Information Processing Systems},
  volume~34, pp.\  12454--12465, 2021.

\bibitem[Karras et~al.(2022)Karras, Aittala, Aila, and Laine]{karras2022edm}
Karras, T., Aittala, M., Aila, T., and Laine, S.
\newblock Elucidating the design space of diffusion-based generative models.
\newblock In \emph{Advances in Neural Information Processing Systems},
  volume~35, 2022.

\bibitem[Kim et~al.(2023)Kim, Lee, and Park]{kim2023stasy}
Kim, J., Lee, C., and Park, N.
\newblock {STaSy}: Score-based tabular data synthesis.
\newblock In \emph{International Conference on Learning Representations}, 2023.

\bibitem[Kipf \& Welling(2017)Kipf and Welling]{kipf2017semi}
Kipf, T.~N. and Welling, M.
\newblock Semi-supervised classification with graph convolutional networks.
\newblock In \emph{International Conference on Learning Representations}, 2017.

\bibitem[Kotelnikov et~al.(2023)Kotelnikov, Baranchuk, Rubachev, and
  Babenko]{kotelnikov2023tabddpm}
Kotelnikov, A., Baranchuk, D., Rubachev, I., and Babenko, A.
\newblock {TabDDPM}: Modelling tabular data with diffusion models.
\newblock In \emph{International Conference on Machine Learning}, volume 202,
  pp.\  17564--17579. PMLR, 2023.

\bibitem[Lee et~al.(2023)Lee, Kim, and Park]{lee2023codi}
Lee, C., Kim, J., and Park, N.
\newblock {CoDi}: Co-evolving contrastive diffusion models for mixed-type
  tabular synthesis.
\newblock In \emph{International Conference on Machine Learning}, volume 202,
  pp.\  18940--18956. PMLR, 2023.

\bibitem[Micci-Barreca(2001)]{miccibarreca2001preprocessing}
Micci-Barreca, D.
\newblock A preprocessing scheme for high-cardinality categorical attributes in
  classification and prediction problems.
\newblock \emph{ACM SIGKDD Explorations Newsletter}, 3\penalty0 (1):\penalty0
  27--32, 2001.

\bibitem[Oh et~al.(2020)Oh, Adamczewski, and Park]{oh2020radial}
Oh, C., Adamczewski, K., and Park, M.
\newblock Radial and directional posteriors for bayesian deep learning.
\newblock In \emph{Proceedings of the AAAI Conference on Artificial
  Intelligence}, volume~34, pp.\  5298--5305, 2020.
\newblock \doi{10.1609/aaai.v34i04.5976}.

\bibitem[Patki et~al.(2016)Patki, Wedge, and Veeramachaneni]{patki2016sdv}
Patki, N., Wedge, R., and Veeramachaneni, K.
\newblock The synthetic data vault.
\newblock In \emph{IEEE International Conference on Data Science and Advanced
  Analytics}, pp.\  399--410, 2016.

\bibitem[Shi et~al.(2025)Shi, Xu, Hua, Zhang, Ermon, and
  Leskovec]{shi2025tabdiff}
Shi, J., Xu, M., Hua, H., Zhang, H., Ermon, S., and Leskovec, J.
\newblock {TabDiff}: a mixed-type diffusion model for tabular data generation.
\newblock In \emph{International Conference on Learning Representations}, 2025.
\newblock arXiv:2410.20626.

\bibitem[Su et~al.(2024)Su, Lu, Pan, Murtadha, Wen, and Liu]{su2024roformer}
Su, J., Lu, Y., Pan, S., Murtadha, A., Wen, B., and Liu, Y.
\newblock {RoFormer}: Enhanced transformer with rotary position embedding.
\newblock \emph{Neurocomputing}, 568:\penalty0 127063, 2024.
\newblock \doi{10.1016/j.neucom.2023.127063}.

\bibitem[Vaswani et~al.(2017)Vaswani, Shazeer, Parmar, Uszkoreit, Jones, Gomez,
  Kaiser, and Polosukhin]{vaswani2017attention}
Vaswani, A., Shazeer, N., Parmar, N., Uszkoreit, J., Jones, L., Gomez, A.~N.,
  Kaiser, {\L}., and Polosukhin, I.
\newblock Attention is all you need.
\newblock In \emph{Advances in Neural Information Processing Systems},
  volume~30, 2017.

\bibitem[Veli{\v{c}}kovi{\'c} et~al.(2018)Veli{\v{c}}kovi{\'c}, Cucurull,
  Casanova, Romero, Li{\`o}, and Bengio]{velickovic2018gat}
Veli{\v{c}}kovi{\'c}, P., Cucurull, G., Casanova, A., Romero, A., Li{\`o}, P.,
  and Bengio, Y.
\newblock Graph attention networks.
\newblock In \emph{International Conference on Learning Representations}, 2018.

\bibitem[Xu et~al.(2019)Xu, Skoularidou, Cuesta-Infante, and
  Veeramachaneni]{xu2019modeling}
Xu, L., Skoularidou, M., Cuesta-Infante, A., and Veeramachaneni, K.
\newblock Modeling tabular data using conditional {GAN}.
\newblock In \emph{Advances in Neural Information Processing Systems},
  volume~32, 2019.

\bibitem[Zhang et~al.(2024)Zhang, Zhang, Srinivasan, Shen, Qin, Faloutsos,
  Rangwala, and Karypis]{zhang2024tabsyn}
Zhang, H., Zhang, J., Srinivasan, B., Shen, Z., Qin, X., Faloutsos, C.,
  Rangwala, H., and Karypis, G.
\newblock Mixed-type tabular data synthesis with score-based diffusion in
  latent space.
\newblock In \emph{International Conference on Learning Representations}, 2024.

\end{thebibliography}
\bibliographystyle{icml2026}

\appendix
\onecolumn
\section{Extended Results}
\label{app:extended}

\subsection{Full Loss Ablation}
\label{app:loss_ablation}

Table~\ref{tab:ablation_full} presents the complete loss ablation across the 10 benchmark datasets for Shape, Trend, and MLE metrics for the MLP architecture.

\subsection{Cross-Architecture Wiring Diagrams}
\label{app:cross_arch_architectures}

Section~\ref{sec:cross_arch_variants} describes how the geometric-input, prediction-head, and augmentation-path mechanisms transfer across the three diffusion denoising backbones evaluated in Section~\ref{sec:main_results}. Figure~\ref{fig:architecture} (main text) shows the Diffusion-MLP variant. This appendix provides the corresponding wiring diagrams for the GNN and Transformer variants and consolidates the behavioral details that apply to all three.

\paragraph{Shared mechanisms across all three diffusion backbones.}
Three mechanisms are common to every $+\textsc{Geom}$ instantiation, regardless of backbone family:

\emph{Geometric inputs are computed from the current state at every forward pass.} At each training step the inputs are derived from the noised data; at each sampling step they are derived from the partially-denoised current state. The input pathway is therefore active throughout reverse diffusion, even though no geometric \emph{supervision} is applied at sampling time (Section~\ref{sec:sampling}).

\emph{Angle head and length head play asymmetric roles.} Both heads attach to the final pre-output representation $\mathbf{h}$ and are supervised during training by $\mathcal{L}_{\text{angle}}$ and $\mathcal{L}_{\text{length}}$. Only the bounded angle prediction $\hat{\boldsymbol{\theta}} = (\pi/2) \tanh(\cdot)$ enters the augmentation path $\mathbf{h}_{\text{aug}} = [\mathbf{h}; \hat{\boldsymbol{\theta}}]$ that feeds the denoising heads (Section~\ref{sec:architecture}, ``Augmented representation''). The length head provides auxiliary supervision only: its forward output is computed at every step but never consumed downstream. Consequently the angle head is essential at inference (the denoising heads cannot run without $\mathbf{h}_{\text{aug}}$), whereas the length head is effectively training-only at the architectural level.

\emph{The denoising heads ($\mathbf{F}_{\text{cont}}$ and per-column categorical logits) always read from $\mathbf{h}_{\text{aug}}$, never from $\mathbf{h}$ directly.} This is what makes geometric supervision an inductive bias on the denoising path rather than a side-task: the gradient from the denoising loss flows back through $\hat{\boldsymbol{\theta}}$ into the angle head and from there into the backbone.

In the figures that follow, the augmentation path is highlighted in red and the standard (denoising-only) path in grey.

\begin{figure*}[h]
\centering
\includegraphics[width=0.85\textwidth]{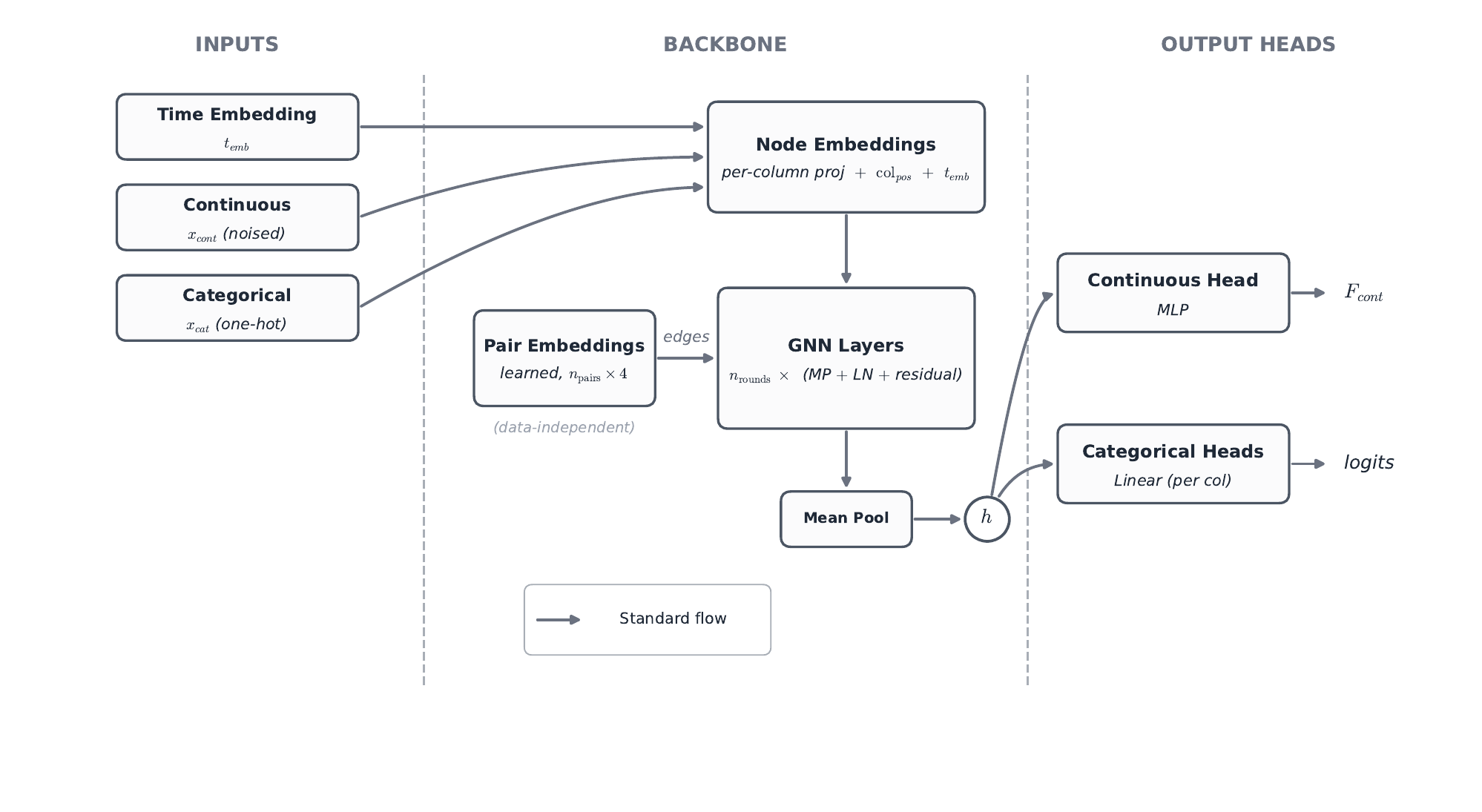}
\caption{\textbf{GNN backbone, no-geometry baseline.} Standard graph denoiser: node embeddings (per-column projection $+$ column-position embedding $+$ time embedding) feed $n_{\text{rounds}}$ rounds of message passing with data-independent learned pair embeddings ($n_{\text{pairs}} \times 4$) as edge attributes; mean-pooled node embeddings feed the denoising heads.}
\label{fig:gnn_nogeom}
\end{figure*}

\subsubsection{Diffusion-MLP Variant}
\label{app:mlp_architecture}

Figure~\ref{fig:architecture} (main text) describes the Diffusion-MLP variant. The $+\textsc{Geom}$ instantiation concatenates the time embedding, noised continuous values, one-hot categoricals, pairwise input angles, and pairwise input lengths into a single flat vector that feeds an input-projection MLP and a stack of $n_{\text{blocks}}$ residual blocks. The NoGeom baseline removes the angle and length input slices, the angle head, the length head, and the augmentation path; the denoising heads then read from $\mathbf{h}$ directly.

\subsubsection{GNN Variant}
\label{app:gnn_architecture}

\begin{figure*}[h]
\centering
\includegraphics[width=0.85\textwidth]{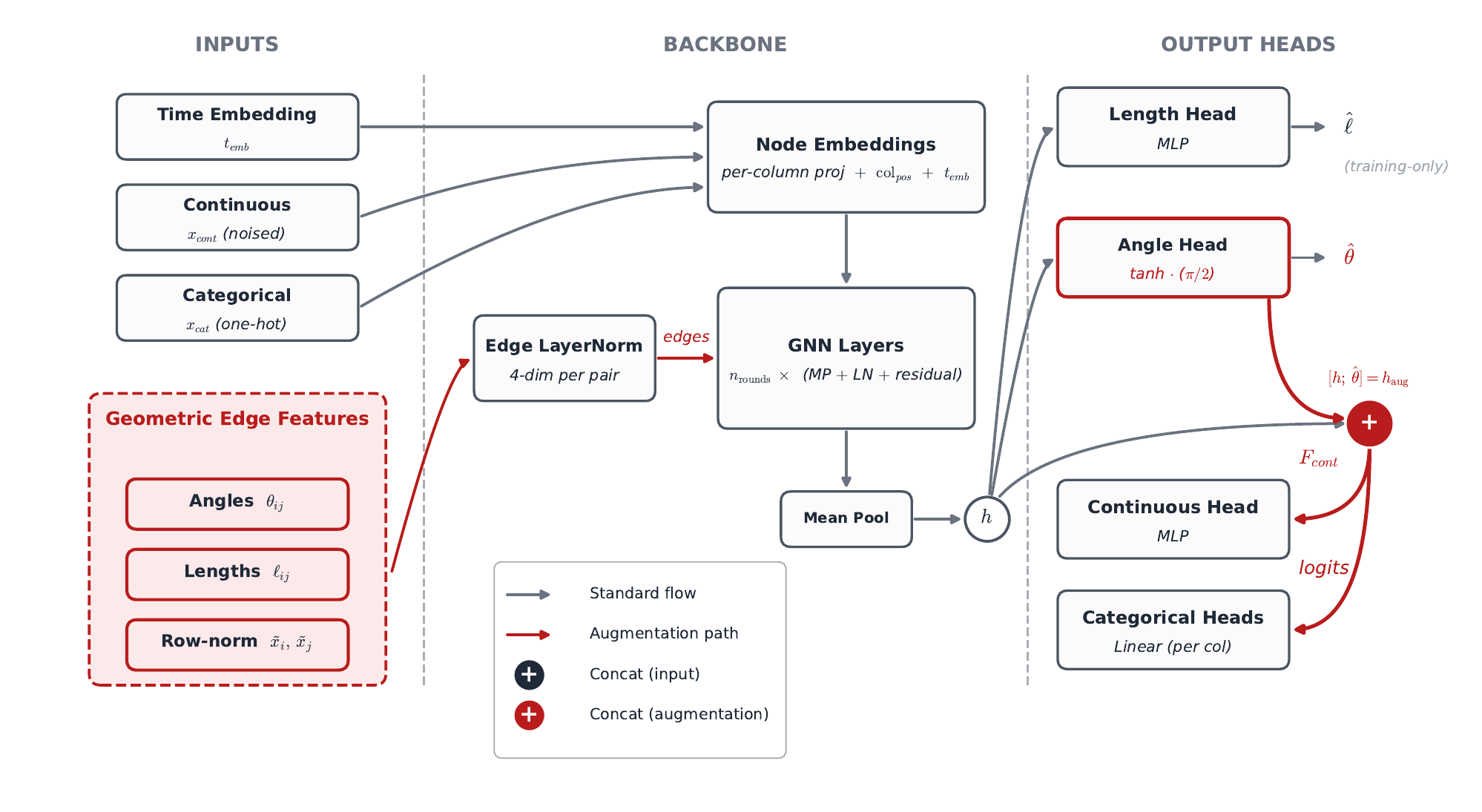}
\caption{\textbf{GNN+Geom architecture.} Four-dimensional geometric edge features (angles $\theta_{ij}$, log-lengths $\ell_{ij}$, row-normalized endpoint values $\tilde{x}_i, \tilde{x}_j$) replace the data-independent learned pair embeddings of the baseline (Figure~\ref{fig:gnn_nogeom}); they pass through Edge LayerNorm into the GNN layers. The augmentation path (red) routes the predicted angle $\hat{\boldsymbol{\theta}}$ through the concat operation $[\mathbf{h}; \hat{\boldsymbol{\theta}}] = \mathbf{h}_{\text{aug}}$ feeding the denoising heads. The length head provides auxiliary training supervision only and is not consumed at inference.}
\label{fig:gnn_geom}
\end{figure*}

The GNN backbone replaces the residual MLP with edge-conditioned message passing on a complete column graph. Nodes correspond to columns and carry a per-column projection of the noised value (continuous: a learned scalar embedding; categorical: a learned embedding of the one-hot vector) summed with a learned column-position embedding and the time embedding broadcast across nodes. The graph is fully connected and a stack of $n_{\text{rounds}}$ message-passing layers (each: edge MLP $\to$ scatter-add aggregation $\to$ node MLP $\to$ residual + LayerNorm) propagates information; the angle direction is sign-flipped for the reverse-direction message to respect the antisymmetry $\theta_{ji} = -\theta_{ij}$. After message passing, node embeddings are mean-pooled into a graph-level vector $\mathbf{h}$ that feeds the prediction heads under the shared scheme described above.

The two variants differ in their edge attributes. The NoGeom baseline (Figure~\ref{fig:gnn_nogeom}) supplies a learned, data-independent embedding per pair, of dimension $n_{\text{pairs}} \times 4$ --- a parameter-matched stand-in for the geometric features that lets the GNN distinguish pairs but provides no information about their current values. The $+\textsc{Geom}$ variant (Figure~\ref{fig:gnn_geom}) replaces these learned embeddings with four data-dependent edge features per pair: the pairwise angle $\theta_{ij}$, the pairwise log-length $\ell_{ij}$, and the row-normalized values $\tilde{x}_i, \tilde{x}_j$ of the two endpoints (centered and scaled by the per-row mean and standard deviation across columns). The four-dimensional edge tensor passes through a LayerNorm before entering the message-passing stack, and the angle and length prediction heads attach to $\mathbf{h}$ in the standard pattern.

\begin{figure*}[h]
\centering
\includegraphics[width=0.85\textwidth]{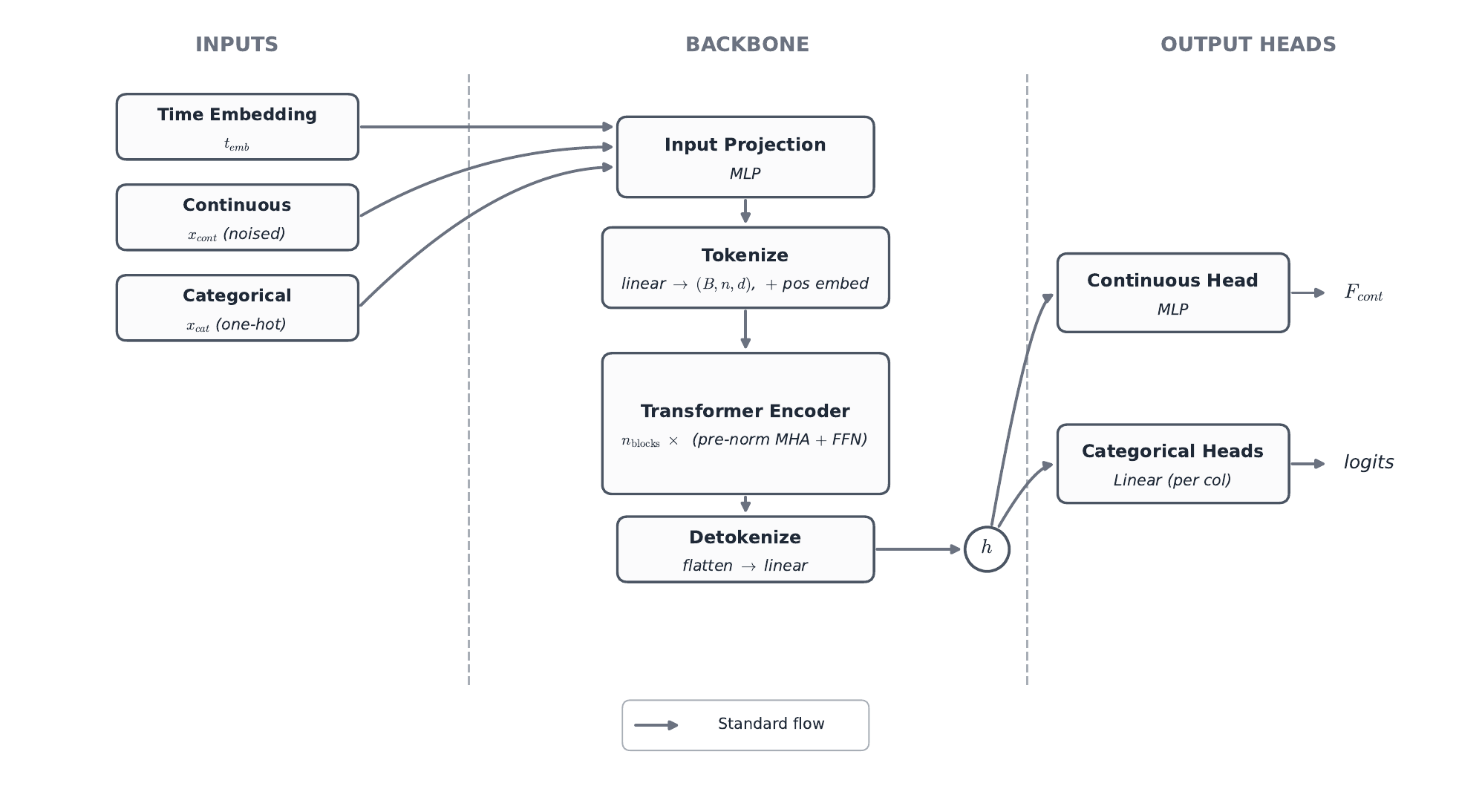}
\caption{\textbf{Transformer backbone, no-geometry baseline.} Time embedding, noised continuous values, and one-hot categoricals concatenate into a flat vector that the input-projection MLP maps to $\mathbf{h}_0$; a linear expansion produces $n_{\text{tokens}}$ latent tokens with learned positional embeddings, processed by a pre-norm Transformer encoder of depth $n_{\text{blocks}}$. A linear contraction flattens the encoded tokens back to a single $d_{\text{model}}$ vector that feeds the denoising heads.}
\label{fig:transformer_nogeom}
\end{figure*}

\begin{figure*}[h]
\centering
\includegraphics[width=0.85\textwidth]{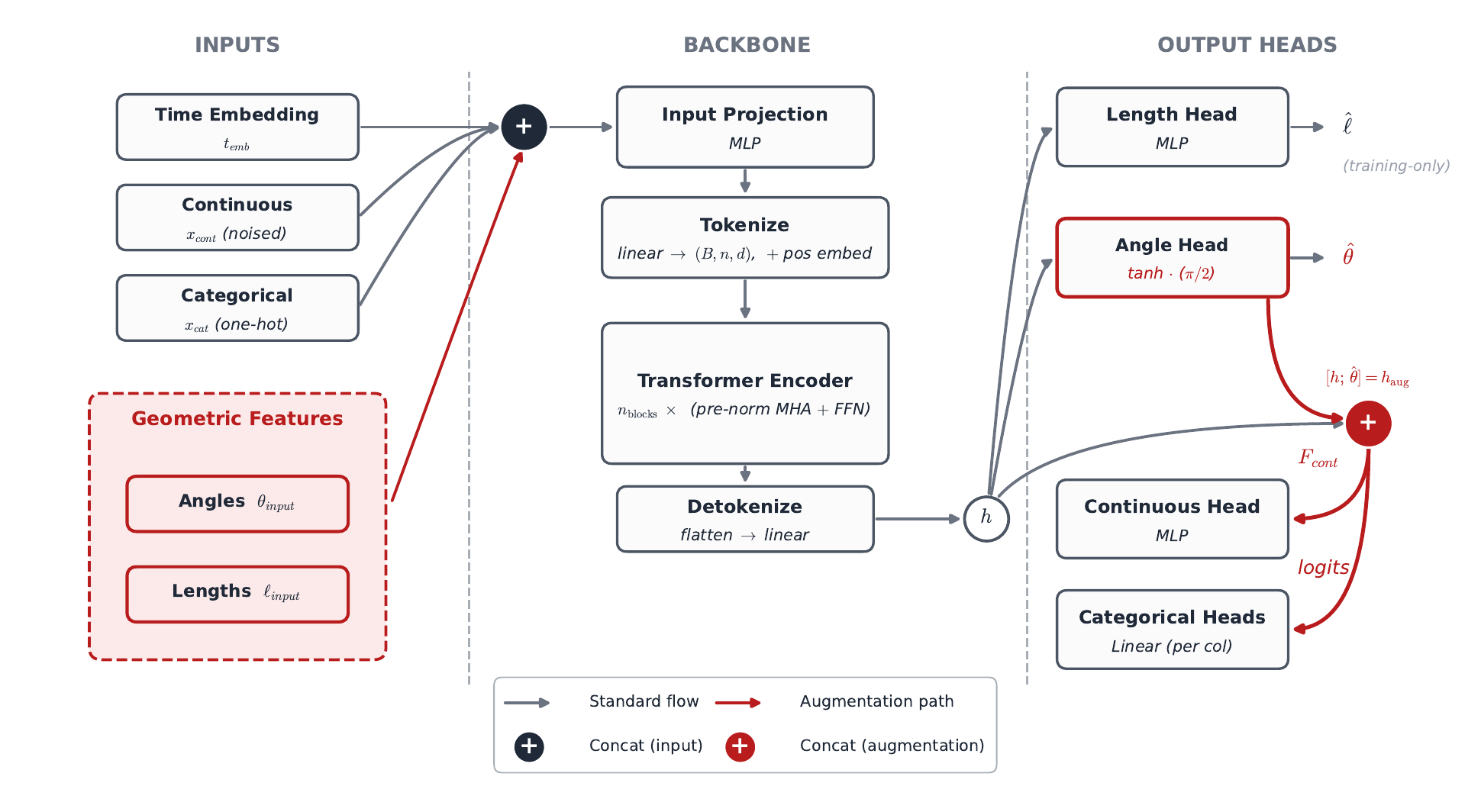}
\caption{\textbf{Transformer+Geom architecture.} Pairwise input angles $\theta_{ij}$ and lengths $\ell_{ij}$ are concatenated alongside the standard inputs before the input-projection MLP; the latent-token Transformer backbone is unchanged from the baseline (Figure~\ref{fig:transformer_nogeom}). The augmentation path (red) routes the predicted angle $\hat{\boldsymbol{\theta}}$ through the concat operation $[\mathbf{h}; \hat{\boldsymbol{\theta}}] = \mathbf{h}_{\text{aug}}$ feeding the denoising heads. The length head provides auxiliary training supervision only.}
\label{fig:transformer_geom}
\end{figure*}

\FloatBarrier

\subsubsection{Transformer Variant}
\label{app:transformer_architecture}

The Transformer variant uses the same flat input vector as the Diffusion-MLP --- time embedding, noised continuous values, one-hot categoricals, and (under $+\textsc{Geom}$) pairwise input angles and lengths concatenated into a single tensor that passes through the shared input-projection MLP to produce a $d_{\text{model}}$-dimensional vector $\mathbf{h}_0$. A linear map then expands $\mathbf{h}_0$ into a sequence of $n_{\text{tokens}}$ latent tokens of dimension $d_{\text{model}}$, a learned positional embedding is added, and a pre-norm Transformer encoder of depth $n_{\text{blocks}}$ with $n_{\text{heads}}$ attention heads processes the token sequence. A final linear map flattens the encoded tokens back to a single $d_{\text{model}}$ vector $\mathbf{h}$ that feeds the prediction heads under the shared scheme. Note that the tokens are not column tokens: they are a fixed-length latent set ($n_{\text{tokens}}=4$ by default) that the encoder uses as a working sequence, with all column information already mixed into $\mathbf{h}_0$ before tokenization.

The NoGeom baseline (Figure~\ref{fig:transformer_nogeom}) removes the angle and length input slices, the angle head, the length head, and the augmentation path. The $+\textsc{Geom}$ variant (Figure~\ref{fig:transformer_geom}) restores all four and routes the predicted angle into the augmentation path in the standard pattern.

\subsection{Architecture Ablation: Effect of Residual Blocks}
\label{app:architecture}

Table~\ref{tab:blocks_comparison} compares 0 vs.\ 8 residual blocks across all datasets. The pattern is consistent: classification tasks favor 0 blocks for fidelity metrics (Shape, Trend), while regression tasks favor 8 blocks for downstream utility (MLE). Exceptions occur precisely in low-categorical datasets (California, Power with 0 categorical columns; Magic with 1), where the per-dataset gain from geometric supervision is smaller on the MLP backbone (Table~\ref{tab:cat_analysis}).

\begin{table*}[!hpb]
\centering
\caption{Ablation study: Effect of residual blocks on performance. Bold indicates better performance. For Shape/Trend, Improv.\ shows improvement from 0 to 8 blocks (\textcolor{blue}{Blue}$\uparrow$ = 8 blocks better, \textcolor{red}{Red}$\downarrow$ = 0 blocks better). For MLE, Average shows \% gap from Real ($\downarrow$ better). Dashes indicate missing data.}
\label{tab:blocks_comparison}
\resizebox{\textwidth}{!}{%
\begin{tabular}{llccccccccccc}
\toprule
Metric & Config & Adult & Default & Diabetes & Magic & Shoppers & Beijing & Bikeshare & California & News & Power & Average \\
\midrule
\multirow{3}{*}{\rotatebox{90}{Shape}}
& 0 Blocks & \textbf{0.493}$_{\pm0.0442}$ & \textbf{0.891}$_{\pm0.0928}$ & \textbf{0.575}$_{\pm0.0286}$ & \textbf{0.814}$_{\pm0.0802}$ & \textbf{0.845}$_{\pm0.0725}$ & 0.956$_{\pm0.0617}$ & 0.830$_{\pm0.0568}$ & \textbf{0.877}$_{\pm0.0795}$ & 2.43$_{\pm0.170}$ & 1.00$_{\pm0.171}$ & \textbf{0.971} \\
& 8 Blocks & 1.03$_{\pm0.0417}$ & 1.02$_{\pm0.0575}$ & 1.04$_{\pm0.0249}$ & 0.967$_{\pm0.0762}$ & 1.15$_{\pm0.0803}$ & \textbf{0.837}$_{\pm0.0466}$ & \textbf{0.745}$_{\pm0.0847}$ & 0.963$_{\pm0.0451}$ & \textbf{1.59}$_{\pm0.0430}$ & \textbf{0.973}$_{\pm0.207}$ & 1.03 \\
& Improv. & \textcolor{red}{109.2\%}$\downarrow$ & \textcolor{red}{14.5\%}$\downarrow$ & \textcolor{red}{81.1\%}$\downarrow$ & \textcolor{red}{18.8\%}$\downarrow$ & \textcolor{red}{36.2\%}$\downarrow$ & \textcolor{blue}{12.5\%}$\uparrow$ & \textcolor{blue}{10.3\%}$\uparrow$ & \textcolor{red}{9.7\%}$\downarrow$ & \textcolor{blue}{34.6\%}$\uparrow$ & \textcolor{blue}{2.7\%}$\uparrow$ & \textcolor{red}{6.3\%}$\downarrow$ \\
\midrule
\multirow{3}{*}{\rotatebox{90}{Trend}}
& 0 Blocks & \textbf{1.23}$_{\pm0.116}$ & \textbf{2.93}$_{\pm0.936}$ & \textbf{1.89}$_{\pm0.146}$ & \textbf{1.10}$_{\pm0.285}$ & \textbf{1.83}$_{\pm0.190}$ & 2.42$_{\pm0.191}$ & 3.12$_{\pm0.376}$ & \textbf{1.28}$_{\pm0.105}$ & 1.53$_{\pm0.190}$ & \textbf{0.335}$_{\pm0.124}$ & \textbf{1.77} \\
& 8 Blocks & 1.98$_{\pm0.227}$ & 3.70$_{\pm1.14}$ & 2.55$_{\pm0.111}$ & 1.36$_{\pm0.323}$ & 1.96$_{\pm0.185}$ & \textbf{2.17}$_{\pm0.221}$ & \textbf{2.86}$_{\pm0.401}$ & 1.33$_{\pm0.403}$ & \textbf{1.20}$_{\pm0.222}$ & 0.387$_{\pm0.147}$ & 1.95 \\
& Improv. & \textcolor{red}{61.2\%}$\downarrow$ & \textcolor{red}{26.6\%}$\downarrow$ & \textcolor{red}{34.7\%}$\downarrow$ & \textcolor{red}{23.6\%}$\downarrow$ & \textcolor{red}{7.3\%}$\downarrow$ & \textcolor{blue}{10.3\%}$\uparrow$ & \textcolor{blue}{8.1\%}$\uparrow$ & \textcolor{red}{3.9\%}$\downarrow$ & \textcolor{blue}{21.4\%}$\uparrow$ & \textcolor{red}{15.3\%}$\downarrow$ & \textcolor{red}{10.5\%}$\downarrow$ \\
\midrule
\multirow{4}{*}{\rotatebox{90}{MLE-1}}
 & & AUC$\uparrow$ & AUC$\uparrow$ & AUC$\uparrow$ & AUC$\uparrow$ & AUC$\uparrow$ & R$^2\uparrow$ & R$^2\uparrow$ & R$^2\uparrow$ & R$^2\uparrow$ & R$^2\uparrow$ & \\
& Real & 0.927$_{\pm0.000543}$ & 0.770$_{\pm0.00214}$ & 0.704$_{\pm0.00161}$ & 0.946$_{\pm0.00169}$ & 0.926$_{\pm0.00260}$ & 0.816$_{\pm0.00720}$ & 0.947$_{\pm0.00254}$ & 0.854$_{\pm0.00334}$ & 0.0921$_{\pm0.0329}$ & 0.938$_{\pm0.0674}$ & 0\% \\
& 0 Blocks & \textbf{0.911}$_{\pm0.00113}$ & \textbf{0.764}$_{\pm0.00373}$ & \textbf{0.694}$_{\pm0.00188}$ & 0.936$_{\pm0.00297}$ & \textbf{0.921}$_{\pm0.00503}$ & 0.665$_{\pm0.0181}$ & 0.930$_{\pm0.00447}$ & 0.816$_{\pm0.00994}$ & 0.0373$_{\pm0.0395}$ & 0.920$_{\pm0.0570}$ & 9.2\% \\
& 8 Blocks & 0.908$_{\pm0.00293}$ & 0.761$_{\pm0.00822}$ & 0.691$_{\pm0.00408}$ & \textbf{0.940}$_{\pm0.00240}$ & 0.915$_{\pm0.00584}$ & \textbf{0.765}$_{\pm0.0103}$ & \textbf{0.939}$_{\pm0.00375}$ & \textbf{0.827}$_{\pm0.00892}$ & \textbf{0.0637}$_{\pm0.0348}$ & \textbf{0.929}$_{\pm0.0600}$ & \textcolor{blue}{4.9\%} \\
\midrule
\multirow{4}{*}{\rotatebox{90}{MLE-2}}
 & & F1$\uparrow$ & F1$\uparrow$ & F1$\uparrow$ & F1$\uparrow$ & F1$\uparrow$ & RMSE$\downarrow$ & RMSE$\downarrow$ & RMSE$\downarrow$ & RMSE$\downarrow$ & RMSE$\downarrow$ & \\
& Real & 0.705$_{\pm0.00548}$ & 0.472$_{\pm0.00696}$ & 0.590$_{\pm0.00319}$ & 0.839$_{\pm0.00676}$ & 0.625$_{\pm0.0124}$ & 0.441$_{\pm0.00869}$ & 0.342$_{\pm0.00824}$ & 0.189$_{\pm0.00216}$ & 0.834$_{\pm0.0153}$ & 0.00855$_{\pm0.00389}$ & 0\% \\
& 0 Blocks & \textbf{0.672}$_{\pm0.00615}$ & 0.451$_{\pm0.0128}$ & \textbf{0.578}$_{\pm0.00443}$ & 0.824$_{\pm0.00665}$ & 0.622$_{\pm0.0181}$ & 0.594$_{\pm0.0159}$ & 0.394$_{\pm0.0125}$ & 0.212$_{\pm0.00565}$ & 0.860$_{\pm0.0177}$ & 0.0103$_{\pm0.00306}$ & 9.9\% \\
& 8 Blocks & 0.664$_{\pm0.00820}$ & \textbf{0.453}$_{\pm0.0109}$ & 0.577$_{\pm0.00485}$ & \textbf{0.831}$_{\pm0.00593}$ & \textbf{0.623}$_{\pm0.0162}$ & \textbf{0.498}$_{\pm0.0109}$ & \textbf{0.368}$_{\pm0.0113}$ & \textbf{0.205}$_{\pm0.00528}$ & \textbf{0.848}$_{\pm0.0147}$ & \textbf{0.00950}$_{\pm0.00335}$ & \textcolor{blue}{5.5\%} \\
\bottomrule
\end{tabular}%
}
\end{table*}

\subsection{Categorical-Anchor Analysis}
\label{app:cat_analysis}

\begin{table}[h]
\centering
\small
\begin{tabular}{lcc}
\toprule
Architecture & Spearman $\rho$ & Spearman $p$ \\
\midrule
MLP & 0.70 & \textbf{0.025} \\
GNN+LE & 0.38 & 0.28 \\
Transformer & 0.42 & 0.22 \\
\midrule
Pooled (30 cells) & 0.51 & \textbf{0.004} \\
\bottomrule
\end{tabular}
\caption{Per-architecture and pooled rank correlation between categorical-column count and per-dataset Shape improvement under $+\textsc{Geom}$ across diffusion denoisers. The original ``categorical anchor'' hypothesis is supported on the MLP track ($\rho = 0.70$, $p = 0.025$) and as a weak aggregate effect across architectures, but does not reach Spearman significance for any individual non-MLP backbone. Trend correlations are not significant for any architecture.}
\label{tab:cat_analysis}
\end{table}

When controlling for total dataset dimensionality by using categorical \emph{fraction} (cat / total cols) instead of raw count, the MLP-track correlation weakens to $\rho = 0.49$ ($p = 0.15$), indicating that the original observation partly conflated ``more categorical columns'' with ``larger overall dataset.'' Continuous-only datasets such as Powerplant (0 categorical columns, $d = 5$) show Shape and Trend gains across multiple diffusion backbones, with the per-cell BH-FDR-corrected effect surviving at $q < 0.05$ for the evaluated diffusion architectures on Trend. We characterize the categorical-anchor pattern as one operating regime among several rather than a necessary condition for $+\textsc{Geom}$. Re-computing the same rank correlation across the loss-ablation configurations of Appendix~\ref{app:loss_ablation} (Table~\ref{tab:cat_anchor_supervision}) shows the pattern is supervision-dependent: \emph{Inputs Only} produces no positive correlation, while every supervised configuration reproduces it. \begin{table}[htpb]
\centering
\small
\begin{tabular}{lcc}
\toprule
Loss config. & Spearman $\rho$ & Spearman $p$ \\
\midrule
Inputs Only & -0.63 & 0.051 \\
Angle Only & \textbf{+0.70} & \textbf{0.025} \\
Angle + Length & \textbf{+0.70} & \textbf{0.025} \\
Full & \textbf{+0.70} & \textbf{0.025} \\
\bottomrule
\end{tabular}
\caption{Rank correlation between categorical-column count and per-dataset Shape improvement (vs NoGeom) on the MLP backbone, across loss-ablation configurations. The anchor pattern appears only under direct supervision; geometric inputs and architecture alone (Inputs Only) do not produce it. $n=10$ datasets.}
\label{tab:cat_anchor_supervision}
\end{table}

\subsection{MLP+Geom vs. TabDiff Per-Dataset Results}
\label{app:sota_per_dataset}
\begin{table*}[!tphb]
\centering
\caption{MLP+Geom vs.\ TabDiff with mean $\pm$ std per dataset. \textbf{Bold} = better cell. $\Delta$ rows: MLP+Geom relative improvement (\textcolor{blue}{$\uparrow$} better; \textcolor{red}{$\downarrow$} worse). For MLE, rightmost column is mean \% gap to \textit{Real} ($\downarrow$ better).}
\label{tab:combined}
\resizebox{\textwidth}{!}{%
\begin{tabular}{llccccccccccc}
\toprule
Metric & Method & Adult & Default & Diabetes & Magic & Shoppers & Beijing & Bikeshare & California & News & Power & Average \\
\midrule
\multirow{3}{*}{\rotatebox{90}{Shape}}
& TabDiff & 0.665\,{\scriptsize$\pm$0.0712} & 1.29\,{\scriptsize$\pm$0.0777} & 1.42\,{\scriptsize$\pm$0.194} & \textbf{0.811\,{\scriptsize$\pm$0.108}} & 1.34\,{\scriptsize$\pm$0.102} & 1.00\,{\scriptsize$\pm$0.0572} & 1.05\,{\scriptsize$\pm$0.0878} & \textbf{0.784\,{\scriptsize$\pm$0.0820}} & 2.41\,{\scriptsize$\pm$0.0448} & 1.11\,{\scriptsize$\pm$0.172} & 1.19 \\
& MLP+Geom & \textbf{0.493\,{\scriptsize$\pm$0.0442}} & \textbf{0.891\,{\scriptsize$\pm$0.0928}} & \textbf{0.472\,{\scriptsize$\pm$0.0208}} & 0.814\,{\scriptsize$\pm$0.0802} & \textbf{0.845\,{\scriptsize$\pm$0.0725}} & \textbf{0.837\,{\scriptsize$\pm$0.0466}} & \textbf{0.745\,{\scriptsize$\pm$0.0847}} & 0.963\,{\scriptsize$\pm$0.0451} & \textbf{1.59\,{\scriptsize$\pm$0.0430}} & \textbf{0.973\,{\scriptsize$\pm$0.207}} & \textbf{0.862} \\
& $\Delta$ & \textcolor{blue}{25.9\%}$\uparrow$ & \textcolor{blue}{30.9\%}$\uparrow$ & \textcolor{blue}{66.6\%}$\uparrow$ & \textcolor{red}{0.4\%}$\downarrow$ & \textcolor{blue}{36.9\%}$\uparrow$ & \textcolor{blue}{16.4\%}$\uparrow$ & \textcolor{blue}{28.8\%}$\uparrow$ & \textcolor{red}{22.8\%}$\downarrow$ & \textcolor{blue}{34.0\%}$\uparrow$ & \textcolor{blue}{12.2\%}$\uparrow$ & \textcolor{blue}{27.3\%}$\uparrow$ \\
\midrule
\multirow{3}{*}{\rotatebox{90}{Trend}}
& TabDiff & 1.53\,{\scriptsize$\pm$0.208} & 4.71\,{\scriptsize$\pm$1.58} & 3.48\,{\scriptsize$\pm$0.516} & \textbf{0.876\,{\scriptsize$\pm$0.228}} & 1.87\,{\scriptsize$\pm$0.127} & 2.60\,{\scriptsize$\pm$0.0933} & 2.90\,{\scriptsize$\pm$0.417} & \textbf{0.547\,{\scriptsize$\pm$0.135}} & 1.74\,{\scriptsize$\pm$0.0697} & \textbf{0.219\,{\scriptsize$\pm$0.0767}} & 2.05 \\
& MLP+Geom & \textbf{1.23\,{\scriptsize$\pm$0.116}} & \textbf{2.93\,{\scriptsize$\pm$0.936}} & \textbf{1.43\,{\scriptsize$\pm$0.0562}} & 1.10\,{\scriptsize$\pm$0.285} & \textbf{1.83\,{\scriptsize$\pm$0.190}} & \textbf{2.17\,{\scriptsize$\pm$0.221}} & \textbf{2.86\,{\scriptsize$\pm$0.401}} & 1.33\,{\scriptsize$\pm$0.403} & \textbf{1.20\,{\scriptsize$\pm$0.222}} & 0.387\,{\scriptsize$\pm$0.147} & \textbf{1.65} \\
& $\Delta$ & \textcolor{blue}{19.6\%}$\uparrow$ & \textcolor{blue}{37.9\%}$\uparrow$ & \textcolor{blue}{58.9\%}$\uparrow$ & \textcolor{red}{25.4\%}$\downarrow$ & \textcolor{blue}{2.3\%}$\uparrow$ & \textcolor{blue}{16.5\%}$\uparrow$ & \textcolor{blue}{1.4\%}$\uparrow$ & \textcolor{red}{143.8\%}$\downarrow$ & \textcolor{blue}{30.9\%}$\uparrow$ & \textcolor{red}{76.3\%}$\downarrow$ & \textcolor{blue}{19.6\%}$\uparrow$ \\
\midrule
\multirow{3}{*}{\rotatebox{90}{MLE-1}}
 & & {\footnotesize\itshape AUC$\uparrow$} & {\footnotesize\itshape AUC$\uparrow$} & {\footnotesize\itshape AUC$\uparrow$} & {\footnotesize\itshape AUC$\uparrow$} & {\footnotesize\itshape AUC$\uparrow$} & {\footnotesize\itshape R$^2\uparrow$} & {\footnotesize\itshape R$^2\uparrow$} & {\footnotesize\itshape R$^2\uparrow$} & {\footnotesize\itshape R$^2\uparrow$} & {\footnotesize\itshape R$^2\uparrow$} & {\footnotesize\itshape gap$\downarrow$} \\
& TabDiff & \textbf{0.912\,{\scriptsize$\pm$0.00211}} & 0.758\,{\scriptsize$\pm$0.00701} & 0.663\,{\scriptsize$\pm$0.0401} & \textbf{0.936\,{\scriptsize$\pm$0.00322}} & \textbf{0.921\,{\scriptsize$\pm$0.00391}} & 0.698\,{\scriptsize$\pm$0.0141} & 0.931\,{\scriptsize$\pm$0.00509} & \textbf{0.829\,{\scriptsize$\pm$0.00828}} & 0.0151\,{\scriptsize$\pm$0.0537} & \textbf{0.931\,{\scriptsize$\pm$0.0330}} & 11.4\% \\
& MLP+Geom & 0.911\,{\scriptsize$\pm$0.00113} & \textbf{0.764\,{\scriptsize$\pm$0.00373}} & \textbf{0.692\,{\scriptsize$\pm$0.00382}} & \textbf{0.936\,{\scriptsize$\pm$0.00297}} & \textbf{0.921\,{\scriptsize$\pm$0.00503}} & \textbf{0.765\,{\scriptsize$\pm$0.0103}} & \textbf{0.939\,{\scriptsize$\pm$0.00375}} & 0.827\,{\scriptsize$\pm$0.00892} & \textbf{0.0637\,{\scriptsize$\pm$0.0348}} & 0.929\,{\scriptsize$\pm$0.0600} & \textcolor{blue}{\textbf{4.8\%}} \\
\midrule
\multirow{3}{*}{\rotatebox{90}{MLE-2}}
 & & {\footnotesize\itshape F1$\uparrow$} & {\footnotesize\itshape F1$\uparrow$} & {\footnotesize\itshape F1$\uparrow$} & {\footnotesize\itshape F1$\uparrow$} & {\footnotesize\itshape F1$\uparrow$} & {\footnotesize\itshape RMSE$\downarrow$} & {\footnotesize\itshape RMSE$\downarrow$} & {\footnotesize\itshape RMSE$\downarrow$} & {\footnotesize\itshape RMSE$\downarrow$} & {\footnotesize\itshape RMSE$\downarrow$} & {\footnotesize\itshape gap$\downarrow$} \\
& TabDiff & 0.667\,{\scriptsize$\pm$0.00854} & 0.439\,{\scriptsize$\pm$0.0148} & 0.525\,{\scriptsize$\pm$0.0527} & 0.818\,{\scriptsize$\pm$0.00784} & 0.617\,{\scriptsize$\pm$0.0171} & 0.564\,{\scriptsize$\pm$0.0131} & 0.393\,{\scriptsize$\pm$0.0143} & \textbf{0.204\,{\scriptsize$\pm$0.00492}} & 0.870\,{\scriptsize$\pm$0.0237} & 0.00978\,{\scriptsize$\pm$0.00185} & 9.7\% \\
& MLP+Geom & \textbf{0.672\,{\scriptsize$\pm$0.00615}} & \textbf{0.451\,{\scriptsize$\pm$0.0128}} & \textbf{0.574\,{\scriptsize$\pm$0.00518}} & \textbf{0.824\,{\scriptsize$\pm$0.00665}} & \textbf{0.622\,{\scriptsize$\pm$0.0181}} & \textbf{0.498\,{\scriptsize$\pm$0.0109}} & \textbf{0.368\,{\scriptsize$\pm$0.0113}} & 0.205\,{\scriptsize$\pm$0.00528} & \textbf{0.848\,{\scriptsize$\pm$0.0147}} & \textbf{0.00950\,{\scriptsize$\pm$0.00335}} & \textcolor{blue}{\textbf{5.6\%}} \\
\bottomrule
\end{tabular}%
}
\end{table*}

\subsection{Ablation: Loss-Component Decomposition}
\label{sec:ablation_loss}

Table~\ref{tab:geometry_ablation_aggregate} dissects the relative contribution of the three geometric loss components on the MLP backbone. Four configurations: (1) \emph{Inputs Only}---geometric features as input but $\lambda_\theta = \lambda_\ell = \lambda_c = 0$ (the architecture-matched supervision-removed configuration analyzed in Section~\ref{sec:capacity_control}); (2) \emph{$\ell$ Loss}---length loss only ($\lambda_\ell=15$, others 0); (3) \emph{$\theta$ Loss}---angle loss only ($\lambda_\theta=15$, others 0); (4) \emph{$\theta + \ell$ Loss}---angle and length ($\lambda_\theta=\lambda_\ell=15$, $\lambda_c=0$); (5) \emph{Full}---complete method ($15, 15, 8$).

\begin{table}[b]
\centering
\small
\caption{Aggregate performance comparison of geometry loss-weight ablations on the MLP backbone. Values are averaged across 3 training seeds, 20 sampling seeds, and all 10 datasets. Gap shows \% distance from Real baseline ($\downarrow$ better).}
\label{tab:geometry_ablation_aggregate}
\begin{tabular}{lcccc}
\toprule
Method & S Err. $\downarrow$ & T Err. $\downarrow$ & MLE-1 $\downarrow$ & MLE-2 $\downarrow$ \\
\midrule
Inputs Only & 1.28 & 1.96 & 7.4\% & 5.6\% \\
$\theta$ Loss & 0.873 & 1.64 & 5.2\% & 5.6\% \\
$\theta + \ell$ Loss & 0.861 & 1.64 & 6.0\% & 6.1\% \\
$\theta + \ell + c$ Loss & 0.862 & 1.65 & 4.8\% & 5.6\% \\
\bottomrule
\end{tabular}
\end{table}

We note an interesting tradeoff between fidelity (Shape and Trend), and utility (MLE). Length-only appears to provide the greatest aggregate utility without much fidelity improvement, while the angle loss alone provides nearly all of the fidelity improvement at the cost of the gains afforded by Length-only utility improvement: the jump from Inputs Only to $\theta$ Loss accounts for 97\% of the Shape improvement and 103\% of the Trend improvement. Adding the length loss further improves Shape (0.873$\to$0.861) but degrades MLE (5.2\%$\to$6.0\%); the consistency loss recovers MLE (6.0\%$\to$4.8\%) while maintaining Shape gains. Figure~\ref{fig:loss_ablation_violin} visualizes the per-dataset trajectory across the five configurations; the MLE-1 panel reveals News as the dominant outlier carrying most of the consistency-loss benefit, a single-dataset effect the aggregate column would otherwise hide. The Inputs Only row complements the architecture-matched analysis in Section~\ref{sec:capacity_control}: even with the geometric heads instantiated and the augmentation path active, gradient descent against the denoising loss alone fails to discover any useful representation of the geometric features.

\FloatBarrier

\begin{table*}[htbp]
\centering
\setlength{\tabcolsep}{4pt}
\renewcommand{\arraystretch}{1.1}
\caption{Loss ablation across all 10 datasets, MLP backbone, 3 training seeds $\times$ 20 generation seeds per cell. \textsc{NoGeom} = no geometric architecture or supervision; \textsc{Inputs Only} = geometric architecture (input concatenation, prediction heads, augmentation path) with all loss weights zeroed ($\lambda_\theta=\lambda_\ell=\lambda_c=0$); \textsc{Angle Only} = $\lambda_\theta=15$, $\lambda_\ell=\lambda_c=0$; \textsc{Length Only} = $\lambda_\ell=15$, $\lambda_\theta=\lambda_c=0$; \textsc{Angle + Length} = $\lambda_\theta=\lambda_\ell=15$, $\lambda_c=0$; \textsc{Full} = $\lambda_\theta=\lambda_\ell=15$, $\lambda_c=8$. \textbf{Bold} marks the best cell per column at displayed precision (ties retained). For MLE-1 the per-dataset metric is AUROC for classification and R$^2$ for regression; for MLE-2 it is F1 and RMSE. The Avg.\ column is the mean error for Shape/Trend and the mean \% gap to \textit{Real} for MLE. Per-dataset relative improvements vs.\ \textsc{NoGeom} are visualized in Figure~\ref{fig:loss_ablation_violin}.}
\label{tab:ablation_full}
\resizebox{\textwidth}{!}{%
\begin{tabular}{llccccccccccc}
\toprule
\textbf{Metric} & \textbf{Config} & \textbf{Adult} & \textbf{Default} & \textbf{Diabetes} & \textbf{Magic} & \textbf{Shoppers} & \textbf{Beijing} & \textbf{Bikeshare} & \textbf{California} & \textbf{News} & \textbf{Power} & \textbf{Avg.} \\
\midrule
\multirow{6}{*}{\rotatebox[origin=c]{90}{\textbf{Shape}}} & NoGeom & 0.819\,{\scriptsize$\pm$0.0454} & 1.17\,{\scriptsize$\pm$0.0780} & 1.27\,{\scriptsize$\pm$0.0210} & 0.943\,{\scriptsize$\pm$0.107} & 1.58\,{\scriptsize$\pm$0.0821} & 0.797\,{\scriptsize$\pm$0.0424} & 0.778\,{\scriptsize$\pm$0.0734} & 1.00\,{\scriptsize$\pm$0.0597} & 2.79\,{\scriptsize$\pm$0.0518} & 1.15\,{\scriptsize$\pm$0.173} & 1.23 \\
& Inputs Only & 0.936\,{\scriptsize$\pm$0.0535} & 1.16\,{\scriptsize$\pm$0.0664} & 1.29\,{\scriptsize$\pm$0.0263} & 0.904\,{\scriptsize$\pm$0.0768} & 1.79\,{\scriptsize$\pm$0.0858} & 0.763\,{\scriptsize$\pm$0.0411} & 0.805\,{\scriptsize$\pm$0.0794} & \textbf{0.959\,{\scriptsize$\pm$0.0536}} & 3.06\,{\scriptsize$\pm$0.156} & 1.13\,{\scriptsize$\pm$0.168} & 1.28 \\
& Angle Only & \textbf{0.484\,{\scriptsize$\pm$0.0327}} & 0.921\,{\scriptsize$\pm$0.0809} & 0.481\,{\scriptsize$\pm$0.0231} & 0.821\,{\scriptsize$\pm$0.0822} & 0.848\,{\scriptsize$\pm$0.0795} & 0.834\,{\scriptsize$\pm$0.0489} & 0.744\,{\scriptsize$\pm$0.0690} & 1.01\,{\scriptsize$\pm$0.0566} & 1.61\,{\scriptsize$\pm$0.0434} & 0.978\,{\scriptsize$\pm$0.196} & 0.873 \\
& Length Only & 0.861\,{\scriptsize$\pm$0.0378} & 1.06\,{\scriptsize$\pm$0.0631} & 1.14\,{\scriptsize$\pm$0.0384} & 1.02\,{\scriptsize$\pm$0.0698} & 1.25\,{\scriptsize$\pm$0.0758} & \textbf{0.686\,{\scriptsize$\pm$0.0360}} & \textbf{0.692\,{\scriptsize$\pm$0.0818}} & 1.13\,{\scriptsize$\pm$0.0520} & 3.15\,{\scriptsize$\pm$0.0362} & 1.01\,{\scriptsize$\pm$0.182} & 1.20 \\
& Angle + Length & 0.491\,{\scriptsize$\pm$0.0438} & \textbf{0.849\,{\scriptsize$\pm$0.0753}} & \textbf{0.464\,{\scriptsize$\pm$0.0207}} & 0.828\,{\scriptsize$\pm$0.0860} & 0.887\,{\scriptsize$\pm$0.109} & 0.845\,{\scriptsize$\pm$0.0401} & 0.742\,{\scriptsize$\pm$0.0881} & 0.960\,{\scriptsize$\pm$0.0492} & \textbf{1.57\,{\scriptsize$\pm$0.0494}} & 0.974\,{\scriptsize$\pm$0.206} & \textbf{0.861} \\
& Full & 0.493\,{\scriptsize$\pm$0.0442} & 0.891\,{\scriptsize$\pm$0.0928} & 0.472\,{\scriptsize$\pm$0.0208} & \textbf{0.814\,{\scriptsize$\pm$0.0802}} & \textbf{0.845\,{\scriptsize$\pm$0.0725}} & 0.837\,{\scriptsize$\pm$0.0466} & 0.745\,{\scriptsize$\pm$0.0847} & 0.963\,{\scriptsize$\pm$0.0451} & 1.59\,{\scriptsize$\pm$0.0430} & \textbf{0.973\,{\scriptsize$\pm$0.207}} & 0.862 \\
\midrule
\multirow{6}{*}{\rotatebox[origin=c]{90}{\textbf{Trend}}} & NoGeom & 1.75\,{\scriptsize$\pm$0.169} & 3.04\,{\scriptsize$\pm$0.774} & 2.48\,{\scriptsize$\pm$0.0593} & \textbf{1.08\,{\scriptsize$\pm$0.308}} & 2.23\,{\scriptsize$\pm$0.200} & 2.10\,{\scriptsize$\pm$0.229} & 2.99\,{\scriptsize$\pm$0.397} & 1.34\,{\scriptsize$\pm$0.172} & 1.57\,{\scriptsize$\pm$0.210} & 0.886\,{\scriptsize$\pm$0.212} & 1.95 \\
& Inputs Only & 1.89\,{\scriptsize$\pm$0.173} & 2.94\,{\scriptsize$\pm$0.798} & 2.60\,{\scriptsize$\pm$0.0545} & 1.14\,{\scriptsize$\pm$0.301} & 2.27\,{\scriptsize$\pm$0.220} & \textbf{1.98\,{\scriptsize$\pm$0.236}} & 3.01\,{\scriptsize$\pm$0.347} & 1.33\,{\scriptsize$\pm$0.203} & 1.58\,{\scriptsize$\pm$0.179} & 0.890\,{\scriptsize$\pm$0.213} & 1.96 \\
& Angle Only & 1.25\,{\scriptsize$\pm$0.130} & \textbf{2.67\,{\scriptsize$\pm$0.827}} & 1.48\,{\scriptsize$\pm$0.0587} & \textbf{1.08\,{\scriptsize$\pm$0.275}} & 1.85\,{\scriptsize$\pm$0.212} & 2.10\,{\scriptsize$\pm$0.210} & 2.94\,{\scriptsize$\pm$0.416} & 1.32\,{\scriptsize$\pm$0.403} & 1.27\,{\scriptsize$\pm$0.251} & 0.440\,{\scriptsize$\pm$0.162} & \textbf{1.64} \\
& Length Only & 1.77\,{\scriptsize$\pm$0.198} & 3.09\,{\scriptsize$\pm$0.833} & 2.45\,{\scriptsize$\pm$0.0820} & 1.29\,{\scriptsize$\pm$0.399} & 1.97\,{\scriptsize$\pm$0.182} & 2.01\,{\scriptsize$\pm$0.259} & 2.81\,{\scriptsize$\pm$0.421} & \textbf{1.15\,{\scriptsize$\pm$0.274}} & 1.66\,{\scriptsize$\pm$0.249} & \textbf{0.357\,{\scriptsize$\pm$0.121}} & 1.86 \\
& Angle + Length & 1.31\,{\scriptsize$\pm$0.204} & 2.87\,{\scriptsize$\pm$0.773} & \textbf{1.43\,{\scriptsize$\pm$0.0479}} & \textbf{1.08\,{\scriptsize$\pm$0.253}} & 1.86\,{\scriptsize$\pm$0.168} & 2.14\,{\scriptsize$\pm$0.248} & \textbf{2.79\,{\scriptsize$\pm$0.365}} & 1.27\,{\scriptsize$\pm$0.319} & 1.24\,{\scriptsize$\pm$0.241} & 0.380\,{\scriptsize$\pm$0.153} & \textbf{1.64} \\
& Full & \textbf{1.23\,{\scriptsize$\pm$0.116}} & 2.93\,{\scriptsize$\pm$0.936} & \textbf{1.43\,{\scriptsize$\pm$0.0562}} & 1.10\,{\scriptsize$\pm$0.285} & \textbf{1.83\,{\scriptsize$\pm$0.190}} & 2.17\,{\scriptsize$\pm$0.221} & 2.86\,{\scriptsize$\pm$0.401} & 1.33\,{\scriptsize$\pm$0.403} & \textbf{1.20\,{\scriptsize$\pm$0.222}} & 0.387\,{\scriptsize$\pm$0.147} & 1.65 \\
\midrule
\multirow{7}{*}{\rotatebox[origin=c]{90}{\textbf{MLE-1}}} &  & {\footnotesize\itshape AUC$\uparrow$} & {\footnotesize\itshape AUC$\uparrow$} & {\footnotesize\itshape AUC$\uparrow$} & {\footnotesize\itshape AUC$\uparrow$} & {\footnotesize\itshape AUC$\uparrow$} & {\footnotesize\itshape R$^2\uparrow$} & {\footnotesize\itshape R$^2\uparrow$} & {\footnotesize\itshape R$^2\uparrow$} & {\footnotesize\itshape R$^2\uparrow$} & {\footnotesize\itshape R$^2\uparrow$} & {\footnotesize\itshape gap$\downarrow$} \\
& NoGeom & 0.909\,{\scriptsize$\pm$0.00115} & 0.763\,{\scriptsize$\pm$0.00344} & 0.683\,{\scriptsize$\pm$0.00194} & 0.937\,{\scriptsize$\pm$0.00322} & 0.915\,{\scriptsize$\pm$0.00488} & \textbf{0.770\,{\scriptsize$\pm$0.0117}} & \textbf{0.940\,{\scriptsize$\pm$0.00429}} & 0.828\,{\scriptsize$\pm$0.00756} & 0.0455\,{\scriptsize$\pm$0.0339} & 0.926\,{\scriptsize$\pm$0.0580} & 6.9\% \\
& Inputs Only & 0.909\,{\scriptsize$\pm$0.00162} & 0.763\,{\scriptsize$\pm$0.00372} & 0.690\,{\scriptsize$\pm$0.00206} & \textbf{0.939\,{\scriptsize$\pm$0.00275}} & 0.913\,{\scriptsize$\pm$0.00618} & 0.768\,{\scriptsize$\pm$0.0111} & \textbf{0.940\,{\scriptsize$\pm$0.00383}} & 0.828\,{\scriptsize$\pm$0.00812} & 0.0399\,{\scriptsize$\pm$0.0345} & 0.926\,{\scriptsize$\pm$0.0595} & 7.4\% \\
& Angle Only & \textbf{0.911\,{\scriptsize$\pm$0.00118}} & 0.764\,{\scriptsize$\pm$0.00370} & \textbf{0.692\,{\scriptsize$\pm$0.00206}} & 0.937\,{\scriptsize$\pm$0.00350} & 0.920\,{\scriptsize$\pm$0.00392} & 0.765\,{\scriptsize$\pm$0.0114} & 0.939\,{\scriptsize$\pm$0.00384} & \textbf{0.830\,{\scriptsize$\pm$0.00674}} & 0.0591\,{\scriptsize$\pm$0.0354} & 0.931\,{\scriptsize$\pm$0.0584} & 5.2\% \\
& Length Only & 0.909\,{\scriptsize$\pm$0.00148} & 0.764\,{\scriptsize$\pm$0.00396} & 0.691\,{\scriptsize$\pm$0.00225} & 0.936\,{\scriptsize$\pm$0.00347} & 0.919\,{\scriptsize$\pm$0.00665} & 0.769\,{\scriptsize$\pm$0.0105} & 0.938\,{\scriptsize$\pm$0.00414} & 0.816\,{\scriptsize$\pm$0.00996} & \textbf{0.0781\,{\scriptsize$\pm$0.0172}} & \textbf{0.934\,{\scriptsize$\pm$0.0471}} & \textbf{3.3\%} \\
& Angle + Length & \textbf{0.911\,{\scriptsize$\pm$0.00115}} & \textbf{0.765\,{\scriptsize$\pm$0.00445}} & \textbf{0.692\,{\scriptsize$\pm$0.00290}} & 0.937\,{\scriptsize$\pm$0.00281} & 0.920\,{\scriptsize$\pm$0.00450} & 0.761\,{\scriptsize$\pm$0.0104} & \textbf{0.940\,{\scriptsize$\pm$0.00437}} & 0.829\,{\scriptsize$\pm$0.00778} & 0.0536\,{\scriptsize$\pm$0.0310} & 0.923\,{\scriptsize$\pm$0.0685} & 6.0\% \\
& Full & \textbf{0.911\,{\scriptsize$\pm$0.00113}} & 0.764\,{\scriptsize$\pm$0.00373} & \textbf{0.692\,{\scriptsize$\pm$0.00382}} & 0.936\,{\scriptsize$\pm$0.00297} & \textbf{0.921\,{\scriptsize$\pm$0.00503}} & 0.765\,{\scriptsize$\pm$0.0103} & 0.939\,{\scriptsize$\pm$0.00375} & 0.827\,{\scriptsize$\pm$0.00892} & 0.0637\,{\scriptsize$\pm$0.0348} & 0.929\,{\scriptsize$\pm$0.0600} & 4.8\% \\
\midrule
\multirow{7}{*}{\rotatebox[origin=c]{90}{\textbf{MLE-2}}} &  & {\footnotesize\itshape F1$\uparrow$} & {\footnotesize\itshape F1$\uparrow$} & {\footnotesize\itshape F1$\uparrow$} & {\footnotesize\itshape F1$\uparrow$} & {\footnotesize\itshape F1$\uparrow$} & {\footnotesize\itshape RMSE$\downarrow$} & {\footnotesize\itshape RMSE$\downarrow$} & {\footnotesize\itshape RMSE$\downarrow$} & {\footnotesize\itshape RMSE$\downarrow$} & {\footnotesize\itshape RMSE$\downarrow$} & {\footnotesize\itshape gap$\downarrow$} \\
& NoGeom & 0.670\,{\scriptsize$\pm$0.00545} & \textbf{0.456\,{\scriptsize$\pm$0.0116}} & 0.576\,{\scriptsize$\pm$0.00723} & 0.824\,{\scriptsize$\pm$0.00672} & 0.619\,{\scriptsize$\pm$0.0192} & \textbf{0.493\,{\scriptsize$\pm$0.0125}} & 0.367\,{\scriptsize$\pm$0.0129} & 0.205\,{\scriptsize$\pm$0.00446} & 0.856\,{\scriptsize$\pm$0.0153} & 0.00980\,{\scriptsize$\pm$0.00314} & 5.8\% \\
& Inputs Only & 0.666\,{\scriptsize$\pm$0.00643} & 0.455\,{\scriptsize$\pm$0.0114} & 0.581\,{\scriptsize$\pm$0.00539} & \textbf{0.828\,{\scriptsize$\pm$0.00681}} & \textbf{0.627\,{\scriptsize$\pm$0.0164}} & 0.494\,{\scriptsize$\pm$0.0119} & \textbf{0.364\,{\scriptsize$\pm$0.0117}} & 0.205\,{\scriptsize$\pm$0.00480} & 0.859\,{\scriptsize$\pm$0.0148} & 0.00979\,{\scriptsize$\pm$0.00320} & 5.6\% \\
& Angle Only & 0.671\,{\scriptsize$\pm$0.00591} & 0.451\,{\scriptsize$\pm$0.0131} & 0.573\,{\scriptsize$\pm$0.00529} & 0.824\,{\scriptsize$\pm$0.00752} & 0.618\,{\scriptsize$\pm$0.0194} & 0.497\,{\scriptsize$\pm$0.0121} & 0.370\,{\scriptsize$\pm$0.0115} & \textbf{0.204\,{\scriptsize$\pm$0.00404}} & 0.850\,{\scriptsize$\pm$0.0164} & 0.00939\,{\scriptsize$\pm$0.00323} & 5.6\% \\
& Length Only & 0.670\,{\scriptsize$\pm$0.00541} & \textbf{0.456\,{\scriptsize$\pm$0.0121}} & \textbf{0.582\,{\scriptsize$\pm$0.00476}} & 0.825\,{\scriptsize$\pm$0.00719} & 0.615\,{\scriptsize$\pm$0.0227} & \textbf{0.493\,{\scriptsize$\pm$0.0113}} & 0.370\,{\scriptsize$\pm$0.0124} & 0.212\,{\scriptsize$\pm$0.00571} & \textbf{0.841\,{\scriptsize$\pm$0.00790}} & \textbf{0.00934\,{\scriptsize$\pm$0.00261}} & \textbf{5.5\%} \\
& Angle + Length & \textbf{0.672\,{\scriptsize$\pm$0.00561}} & 0.451\,{\scriptsize$\pm$0.0122} & 0.574\,{\scriptsize$\pm$0.00499} & 0.824\,{\scriptsize$\pm$0.00659} & 0.614\,{\scriptsize$\pm$0.0166} & 0.502\,{\scriptsize$\pm$0.0109} & 0.366\,{\scriptsize$\pm$0.0132} & \textbf{0.204\,{\scriptsize$\pm$0.00464}} & 0.852\,{\scriptsize$\pm$0.0142} & 0.00979\,{\scriptsize$\pm$0.00373} & 6.1\% \\
& Full & \textbf{0.672\,{\scriptsize$\pm$0.00615}} & 0.451\,{\scriptsize$\pm$0.0128} & 0.574\,{\scriptsize$\pm$0.00518} & 0.824\,{\scriptsize$\pm$0.00665} & 0.622\,{\scriptsize$\pm$0.0181} & 0.498\,{\scriptsize$\pm$0.0109} & 0.368\,{\scriptsize$\pm$0.0113} & 0.205\,{\scriptsize$\pm$0.00528} & 0.848\,{\scriptsize$\pm$0.0147} & 0.00950\,{\scriptsize$\pm$0.00335} & 5.6\% \\
\bottomrule
\end{tabular}%
}
\end{table*}

\subsection{Per-Dataset Downstream Utility Across Architectures}
\label{app:mle_per_dataset}

Tables~\ref{tab:mle1_appendix} and \ref{tab:mle2_appendix} report per-dataset MLE-1 (AUROC for classification, R$^2$ for regression) and MLE-2 (F1 for classification, RMSE for regression) for all three diffusion backbone pairs on all 10 datasets. Aggregated cross-architecture statistical tests appear in Section~\ref{sec:main_results}; the supervision-vs-capacity attribution on the MLP track appears in Section~\ref{sec:capacity_control}.

\begin{table*}[t]
\centering
\setlength{\tabcolsep}{4pt}
\renewcommand{\arraystretch}{1.1}
\caption{Per-dataset MLE-1 (AUROC for classification, R$^2$ for regression) for all three backbone pairs. \textbf{Bold} marks the better cell within each pair. The rightmost column reports mean \% gap to \textit{Real} ($\downarrow$ better).}
\label{tab:mle1_appendix}
\resizebox{\textwidth}{!}{%
\begin{tabular}{lccccccccccc}
\toprule
\textbf{Method} & \textbf{Adult} & \textbf{Default} & \textbf{Diabetes} & \textbf{Magic} & \textbf{Shoppers} & \textbf{Beijing} & \textbf{Bikeshare} & \textbf{California} & \textbf{News} & \textbf{Power} & \textbf{Avg.} \\
\midrule
  & {\footnotesize\itshape AUC$\uparrow$} & {\footnotesize\itshape AUC$\uparrow$} & {\footnotesize\itshape AUC$\uparrow$} & {\footnotesize\itshape AUC$\uparrow$} & {\footnotesize\itshape AUC$\uparrow$} & {\footnotesize\itshape R$^2\uparrow$} & {\footnotesize\itshape R$^2\uparrow$} & {\footnotesize\itshape R$^2\uparrow$} & {\footnotesize\itshape R$^2\uparrow$} & {\footnotesize\itshape R$^2\uparrow$} & {\footnotesize\itshape gap$\downarrow$} \\
\textit{Real} & \textit{0.927\,{\scriptsize$\pm$0.000543}} & \textit{0.770\,{\scriptsize$\pm$0.00214}} & \textit{0.704\,{\scriptsize$\pm$0.00161}} & \textit{0.946\,{\scriptsize$\pm$0.00169}} & \textit{0.926\,{\scriptsize$\pm$0.00260}} & \textit{0.816\,{\scriptsize$\pm$0.00720}} & \textit{0.947\,{\scriptsize$\pm$0.00254}} & \textit{0.854\,{\scriptsize$\pm$0.00334}} & \textit{0.0921\,{\scriptsize$\pm$0.0329}} & \textit{0.938\,{\scriptsize$\pm$0.0674}} & \textit{0\%} \\
\cmidrule(lr){2-12}
MLP & 0.909\,{\scriptsize$\pm$0.00115} & 0.763\,{\scriptsize$\pm$0.00344} & 0.683\,{\scriptsize$\pm$0.00194} & \textbf{0.937\,{\scriptsize$\pm$0.00322}} & 0.915\,{\scriptsize$\pm$0.00488} & \textbf{0.770\,{\scriptsize$\pm$0.0117}} & \textbf{0.940\,{\scriptsize$\pm$0.00429}} & \textbf{0.828\,{\scriptsize$\pm$0.00756}} & 0.0455\,{\scriptsize$\pm$0.0339} & 0.926\,{\scriptsize$\pm$0.0580} & 6.9\% \\
MLP+Geom & \textbf{0.911\,{\scriptsize$\pm$0.00113}} & \textbf{0.764\,{\scriptsize$\pm$0.00373}} & \textbf{0.692\,{\scriptsize$\pm$0.00382}} & 0.936\,{\scriptsize$\pm$0.00297} & \textbf{0.921\,{\scriptsize$\pm$0.00503}} & 0.765\,{\scriptsize$\pm$0.0103} & 0.939\,{\scriptsize$\pm$0.00375} & 0.827\,{\scriptsize$\pm$0.00892} & \textbf{0.0637\,{\scriptsize$\pm$0.0348}} & \textbf{0.929\,{\scriptsize$\pm$0.0600}} & \textbf{4.8\%} \\
\cmidrule(lr){2-12}
GNN+LE & 0.909\,{\scriptsize$\pm$0.00148} & \textbf{0.763\,{\scriptsize$\pm$0.00339}} & 0.682\,{\scriptsize$\pm$0.00645} & 0.642\,{\scriptsize$\pm$0.211} & 0.916\,{\scriptsize$\pm$0.00509} & \textbf{0.725\,{\scriptsize$\pm$0.0138}} & \textbf{0.936\,{\scriptsize$\pm$0.00350}} & 0.806\,{\scriptsize$\pm$0.0122} & 0.0217\,{\scriptsize$\pm$0.0373} & 0.917\,{\scriptsize$\pm$0.0567} & 13.6\% \\
GNN+Geom & \textbf{0.911\,{\scriptsize$\pm$0.00109}} & \textbf{0.763\,{\scriptsize$\pm$0.00351}} & \textbf{0.692\,{\scriptsize$\pm$0.00340}} & \textbf{0.934\,{\scriptsize$\pm$0.00308}} & \textbf{0.919\,{\scriptsize$\pm$0.00552}} & 0.685\,{\scriptsize$\pm$0.0170} & 0.930\,{\scriptsize$\pm$0.00366} & \textbf{0.808\,{\scriptsize$\pm$0.0100}} & \textbf{0.0471\,{\scriptsize$\pm$0.0390}} & \textbf{0.920\,{\scriptsize$\pm$0.0528}} & \textbf{8.0\%} \\
\cmidrule(lr){2-12}
\cmidrule(lr){2-12}
Transformer & 0.908\,{\scriptsize$\pm$0.00262} & \textbf{0.763\,{\scriptsize$\pm$0.00473}} & 0.688\,{\scriptsize$\pm$0.00421} & \textbf{0.939\,{\scriptsize$\pm$0.00275}} & 0.917\,{\scriptsize$\pm$0.00596} & \textbf{0.762\,{\scriptsize$\pm$0.0135}} & \textbf{0.939\,{\scriptsize$\pm$0.00424}} & 0.821\,{\scriptsize$\pm$0.00871} & 0.0415\,{\scriptsize$\pm$0.0366} & 0.926\,{\scriptsize$\pm$0.0528} & 7.5\% \\
Transformer+Geom & \textbf{0.911\,{\scriptsize$\pm$0.00135}} & 0.762\,{\scriptsize$\pm$0.00635} & \textbf{0.692\,{\scriptsize$\pm$0.00430}} & 0.938\,{\scriptsize$\pm$0.00280} & \textbf{0.920\,{\scriptsize$\pm$0.00503}} & 0.736\,{\scriptsize$\pm$0.0127} & 0.936\,{\scriptsize$\pm$0.00419} & \textbf{0.827\,{\scriptsize$\pm$0.00873}} & \textbf{0.0437\,{\scriptsize$\pm$0.0396}} & \textbf{0.931\,{\scriptsize$\pm$0.0555}} & \textbf{7.3\%} \\
\bottomrule
\end{tabular}%
}
\end{table*}

\begin{table*}[t]
\centering
\setlength{\tabcolsep}{4pt}
\renewcommand{\arraystretch}{1.1}
\caption{Per-dataset MLE-2 (F1 for classification, RMSE for regression) for all three backbone pairs. \textbf{Bold} marks the better cell within each pair. The rightmost column reports mean \% gap to \textit{Real} ($\downarrow$ better).}
\label{tab:mle2_appendix}
\resizebox{\textwidth}{!}{%
\begin{tabular}{lccccccccccc}
\toprule
\textbf{Method} & \textbf{Adult} & \textbf{Default} & \textbf{Diabetes} & \textbf{Magic} & \textbf{Shoppers} & \textbf{Beijing} & \textbf{Bikeshare} & \textbf{California} & \textbf{News} & \textbf{Power} & \textbf{Avg.} \\
\midrule
  & {\footnotesize\itshape F1$\uparrow$} & {\footnotesize\itshape F1$\uparrow$} & {\footnotesize\itshape F1$\uparrow$} & {\footnotesize\itshape F1$\uparrow$} & {\footnotesize\itshape F1$\uparrow$} & {\footnotesize\itshape RMSE$\downarrow$} & {\footnotesize\itshape RMSE$\downarrow$} & {\footnotesize\itshape RMSE$\downarrow$} & {\footnotesize\itshape RMSE$\downarrow$} & {\footnotesize\itshape RMSE$\downarrow$} & {\footnotesize\itshape gap$\downarrow$} \\
\textit{Real} & \textit{0.705\,{\scriptsize$\pm$0.00548}} & \textit{0.472\,{\scriptsize$\pm$0.00696}} & \textit{0.590\,{\scriptsize$\pm$0.00319}} & \textit{0.839\,{\scriptsize$\pm$0.00676}} & \textit{0.625\,{\scriptsize$\pm$0.0124}} & \textit{0.441\,{\scriptsize$\pm$0.00869}} & \textit{0.342\,{\scriptsize$\pm$0.00824}} & \textit{0.189\,{\scriptsize$\pm$0.00216}} & \textit{0.834\,{\scriptsize$\pm$0.0153}} & \textit{0.00855\,{\scriptsize$\pm$0.00389}} & \textit{0\%} \\
\cmidrule(lr){2-12}
MLP & 0.670\,{\scriptsize$\pm$0.00545} & \textbf{0.456\,{\scriptsize$\pm$0.0116}} & \textbf{0.576\,{\scriptsize$\pm$0.00723}} & \textbf{0.824\,{\scriptsize$\pm$0.00672}} & 0.619\,{\scriptsize$\pm$0.0192} & \textbf{0.493\,{\scriptsize$\pm$0.0125}} & \textbf{0.367\,{\scriptsize$\pm$0.0129}} & \textbf{0.205\,{\scriptsize$\pm$0.00446}} & 0.856\,{\scriptsize$\pm$0.0153} & 0.00980\,{\scriptsize$\pm$0.00314} & 5.8\% \\
MLP+Geom & \textbf{0.672\,{\scriptsize$\pm$0.00615}} & 0.451\,{\scriptsize$\pm$0.0128} & 0.574\,{\scriptsize$\pm$0.00518} & \textbf{0.824\,{\scriptsize$\pm$0.00665}} & \textbf{0.622\,{\scriptsize$\pm$0.0181}} & 0.498\,{\scriptsize$\pm$0.0109} & 0.368\,{\scriptsize$\pm$0.0113} & \textbf{0.205\,{\scriptsize$\pm$0.00528}} & \textbf{0.848\,{\scriptsize$\pm$0.0147}} & \textbf{0.00950\,{\scriptsize$\pm$0.00335}} & \textbf{5.6\%} \\
\cmidrule(lr){2-12}
GNN+LE & 0.671\,{\scriptsize$\pm$0.00567} & 0.455\,{\scriptsize$\pm$0.0117} & 0.567\,{\scriptsize$\pm$0.00909} & 0.452\,{\scriptsize$\pm$0.265} & \textbf{0.618\,{\scriptsize$\pm$0.0172}} & \textbf{0.538\,{\scriptsize$\pm$0.0134}} & \textbf{0.377\,{\scriptsize$\pm$0.0103}} & 0.218\,{\scriptsize$\pm$0.00673} & 0.866\,{\scriptsize$\pm$0.0168} & 0.0104\,{\scriptsize$\pm$0.00307} & 13.3\% \\
GNN+Geom & \textbf{0.673\,{\scriptsize$\pm$0.00633}} & \textbf{0.457\,{\scriptsize$\pm$0.0114}} & \textbf{0.575\,{\scriptsize$\pm$0.00466}} & \textbf{0.822\,{\scriptsize$\pm$0.00778}} & 0.617\,{\scriptsize$\pm$0.0204} & 0.576\,{\scriptsize$\pm$0.0155} & 0.394\,{\scriptsize$\pm$0.0103} & \textbf{0.216\,{\scriptsize$\pm$0.00559}} & \textbf{0.855\,{\scriptsize$\pm$0.0177}} & \textbf{0.0103\,{\scriptsize$\pm$0.00283}} & \textbf{9.7\%} \\
\cmidrule(lr){2-12}
\cmidrule(lr){2-12}
Transformer & 0.667\,{\scriptsize$\pm$0.00804} & \textbf{0.452\,{\scriptsize$\pm$0.0155}} & \textbf{0.578\,{\scriptsize$\pm$0.00659}} & \textbf{0.829\,{\scriptsize$\pm$0.00543}} & 0.622\,{\scriptsize$\pm$0.0195} & \textbf{0.501\,{\scriptsize$\pm$0.0140}} & \textbf{0.369\,{\scriptsize$\pm$0.0127}} & 0.209\,{\scriptsize$\pm$0.00505} & 0.858\,{\scriptsize$\pm$0.0160} & 0.00988\,{\scriptsize$\pm$0.00289} & \textbf{6.4\%} \\
Transformer+Geom & \textbf{0.670\,{\scriptsize$\pm$0.00718}} & 0.450\,{\scriptsize$\pm$0.0118} & 0.576\,{\scriptsize$\pm$0.00466} & \textbf{0.829\,{\scriptsize$\pm$0.00696}} & \textbf{0.623\,{\scriptsize$\pm$0.0175}} & 0.527\,{\scriptsize$\pm$0.0126} & 0.378\,{\scriptsize$\pm$0.0123} & \textbf{0.206\,{\scriptsize$\pm$0.00515}} & \textbf{0.856\,{\scriptsize$\pm$0.0180}} & \textbf{0.00945\,{\scriptsize$\pm$0.00303}} & 6.6\% \\
\bottomrule
\end{tabular}%
}
\end{table*}

\subsection{Training Dynamics}
\label{app:training}

\begin{figure*}[htpb]
\centering
\includegraphics[width=\textwidth]{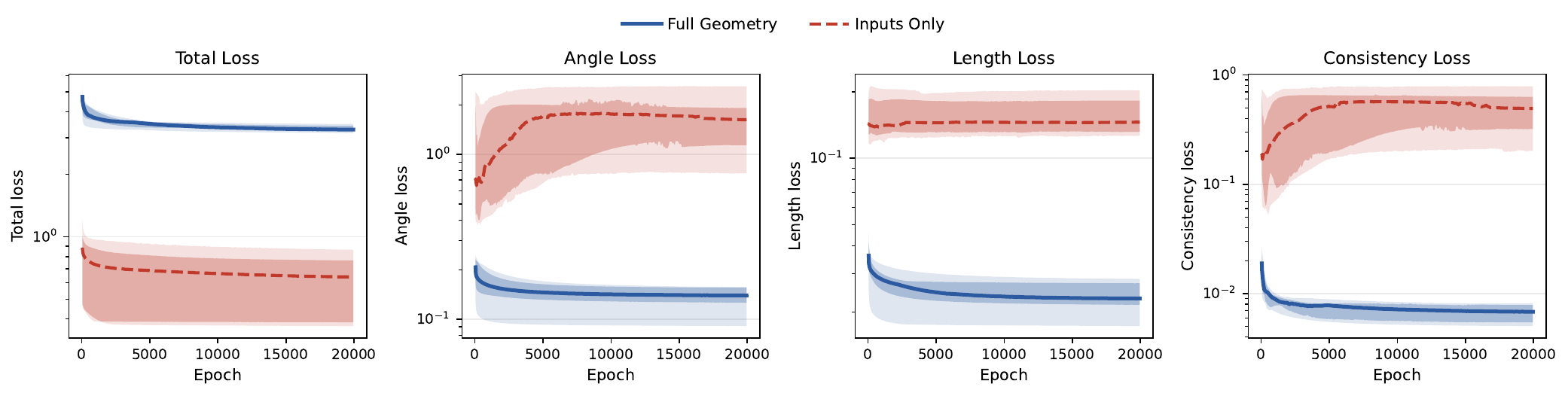}
\caption{\textbf{Training Dynamics (Default).} Full Geometry (solid) vs.\ Inputs Only (dashed). \emph{Total Loss} includes weighted geometric terms ($\lambda_\theta, \lambda_\ell, \lambda_c$ = $15, 15, 8$) for Full Geometry but only diffusion losses ($\mathcal{L}_\text{cont} + \mathcal{L}_\text{cat}$) for Inputs Only. Inputs Only converges to lower total loss (0.60 vs.\ 2.08) yet fails to learn geometry: Angle 2.63 vs.\ 0.07 (37$\times$), Length 0.11 vs.\ 0.01 (8$\times$), Consistency 0.88 vs.\ 0.003 (339$\times$). Weighted geometric loss is 37.4$\times$ higher, showing geometric \emph{inputs} alone provide no learning signal; explicit \emph{supervision} is essential.}
\label{fig:training_curves}
\end{figure*}

Figure~\ref{fig:training_curves} shows training curves comparing Full Geometry vs.\ Inputs Only (geometric inputs without supervision).

\subsection{Detailed Comparison with TabDiff}
\label{app:tabdiff_comparison}

Table~\ref{tab:tabdiff_detailed} provides a comprehensive comparison of shared and differing components between our method and TabDiff.

The shared diffusion framework ensures that performance differences arise from our architectural and supervisory contributions, not from diffusion modifications. The 20$\times$ increase in sampling steps and 2.5$\times$ increase in training epochs represent additional compute; our 3.5$\times$ average parameter reduction partially offsets inference cost. Notably, extended training does not benefit TabDiff: Table~\ref{tab:tabdiff_epochs} shows that TabDiff at 8,000 epochs outperforms or ties TabDiff at 20,000 epochs on 16/24 metrics across 6 datasets, validating our use of TabDiff's published 8,000-epoch protocol.

\begin{table}[h]
\centering
\small
\setlength{\tabcolsep}{5pt}
\begin{tabular}{lcc|lcc}
\toprule
\multicolumn{3}{c|}{\textbf{Shared (Identical)}} &
\multicolumn{3}{c}{\textbf{Different}} \\
\cmidrule(r){1-3} \cmidrule(l){4-6}
Component & TabDiff & Ours &
Component & TabDiff & Ours \\
\midrule
Continuous diffusion & EDM & EDM &
Architecture & Transformer & Residual MLP \\

Categorical diffusion & Masked & Masked &
Column interaction & Self-attention & Geometric features \\

Preconditioning &
$c_{\text{skip}}, c_{\text{out}}, c_{\text{in}}$ &
$c_{\text{skip}}, c_{\text{out}}, c_{\text{in}}$ &
Parameters & $\sim$10M & $\sim$400K--6M \\

Noise schedule & Power-mean & Power-mean &
Geometric supervision & \xmark & \cmark \\

Learnable schedules &
Per-column $\rho$, $k$ &
Per-column $\rho$, $k$ &
Sampling steps & 50 & 1000 \\

$\sigma_{\min}, \sigma_{\max}$ &
0.002, 80 &
0.002, 80 &
Training epochs & 8{,}000 & 20{,}000 \\

Batch size & 4096 & 4096 &
Best model selection & $\geq$4{,}000 & $\geq$10{,}000 \\

& & &
$\lambda_{\text{cat}}$ &
$\sim$1.0 (annealed) &
0.05 (fixed) \\

& & &
$\lambda_{\theta} + \lambda_{\ell} + \lambda_{c}$ &
0 &
38.0 \\
\bottomrule
\end{tabular}
\caption{Detailed comparison with TabDiff. We deliberately share the core diffusion framework (left) to isolate the contribution of geometric features. Key differences (right) are architectural (MLP vs.\ transformer), supervisory (geometric losses), and procedural (more sampling steps and training epochs).}
\label{tab:tabdiff_detailed}
\end{table}

\subsection{TabDiff Extended Training}
\label{app:tabdiff_epochs}

Table~\ref{tab:tabdiff_epochs} compares TabDiff trained for 8,000 epochs (published protocol) versus 20,000 epochs on 6 datasets. Extended training does not improve TabDiff performance---8,000 epochs outperforms or ties 16/24 metrics.

\begin{table*}[h]
\centering
\small
\setlength{\tabcolsep}{6pt}
\begin{tabular}{lcccc}
\toprule
Dataset & Shape (8k / 20k) & Trend (8k / 20k) & AUROC or R$^2$ (8k / 20k) & F1 or RMSE (8k / 20k) \\
\midrule
Adult    & \textbf{0.994} / 0.993 & \textbf{0.986} / 0.985 & 0.912 / \textbf{0.913} & \textbf{0.670} / 0.668 \\
Beijing  & \textbf{0.990} / 0.990 & 0.974 / \textbf{0.975} & \textbf{0.699} / 0.697 & \textbf{0.563} / 0.565 \\
Default  & \textbf{0.987} / 0.987 & \textbf{0.946} / 0.933 & \textbf{0.759} / 0.758 & 0.438 / \textbf{0.438} \\
Magic    & 0.992 / 0.992          & 0.993 / 0.993          & 0.935 / \textbf{0.937} & 0.816 / \textbf{0.822} \\
News     & 0.975 / \textbf{0.977} & 0.982 / \textbf{0.983} & \textbf{0.062} / 0.048 & 0.874 / \textbf{0.872} \\
Shoppers & \textbf{0.987} / 0.986 & \textbf{0.982} / 0.982 & \textbf{0.922} / 0.921 & \textbf{0.616} / 0.612 \\
\midrule
8k wins / 20k wins / ties & 5 / 1 / 0 & 4 / 2 / 0 & 3 / 3 / 0 & 4 / 2 / 0 \\
\bottomrule
\end{tabular}
\caption{TabDiff at 8{,}000 vs.\ 20{,}000 training epochs. Bold = better cell within each (dataset, metric) pair; higher is better for Shape, Trend, AUROC/R$^2$, and F1, while lower is better for RMSE. AUROC/R$^2$ row: AUROC for classification (Adult, Default, Magic, Shoppers), R$^2$ for regression (Beijing, News). F1/RMSE row: F1 for classification, RMSE for regression. Aggregate across the 24 (dataset, metric) cells: 8k wins 16, 20k wins 8. Extended training does not benefit TabDiff, validating our use of their published 8{,}000-epoch protocol. Single training seed, 20 sampling seeds.}
\label{tab:tabdiff_epochs}
\end{table*}

\subsection{Computational Cost}
\label{app:compute}

Table~\ref{tab:compute} compares per-dataset training time, sampling time, and parameter counts between GATD-MLP and TabDiff across all ten benchmark datasets. All measurements are taken on a single NVIDIA T4 GPU under identical batching and I/O conditions.

\paragraph{Analysis.} GATD-MLP achieves substantial parameter reduction (average 0.29$\times$, range 0.04--0.55$\times$) but incurs higher single-dataset sampling cost (average 2.71$\times$) due to 20$\times$ more sampling steps (1000 vs.\ 50). The trade-off varies by dataset: Magic achieves both smaller model \emph{and} faster sampling (0.69$\times$), while regression datasets with $n_{\text{blocks}}=8$ (Beijing, News, Bikesharing, California, Powerplant) have larger models and slower sampling. Classification datasets with $n_{\text{blocks}}=0$ (Adult, Default, Diabetes, Magic, Shoppers) maintain the strongest parameter advantage. Despite 2.5$\times$ more training epochs, total end-to-end training time averages 1.7$\times$ faster than TabDiff due to smaller models and simpler architectures; this is a complete-run wall-clock comparison, not a per-epoch throughput claim.

\begin{table*}[h]
\centering
\small
\begin{tabular}{l|rrr|rrr|rr}
\toprule
& \multicolumn{3}{c|}{\textbf{Geometry (Ours)}} & \multicolumn{3}{c|}{\textbf{TabDiff}} & \multicolumn{2}{c}{\textbf{Ratio (Ours/TabDiff)}} \\
Dataset & Params & Train (h) & Sample (s) & Params & Train (h) & Sample (s) & Params & Sample \\
\midrule
Adult & 542K & 1.25 & 47.8 & 10.6M & 3.62 & 34.6 & 0.05$\times$ & 1.38$\times$ \\
Beijing & 4.67M & 3.18 & 135.7 & 10.6M & 4.21 & 33.0 & 0.44$\times$ & 4.11$\times$ \\
Default & 736K & 1.18 & 34.3 & 10.7M & 3.74 & 29.5 & 0.07$\times$ & 1.16$\times$ \\
Diabetes & 3.97M & 10.25 & 506.7 & 10.8M & 13.51 & 299.5 & 0.37$\times$ & 1.69$\times$ \\
Magic & 401K & 0.37 & 10.1 & 10.6M & 2.24 & 14.6 & 0.04$\times$ & 0.69$\times$ \\
News & 5.96M & 4.34 & 175.7 & 10.9M & 5.34 & 51.4 & 0.55$\times$ & 3.42$\times$ \\
Shoppers & 605K & 0.38 & 20.4 & 10.6M & 2.19 & 19.2 & 0.06$\times$ & 1.07$\times$ \\
\midrule
Bikesharing & 4.68M & 1.14 & 63.9 & 10.6M & 2.48 & 16.3 & 0.44$\times$ & 3.93$\times$ \\
California & 4.58M & 1.32 & 62.3 & 10.6M & 2.43 & 13.8 & 0.43$\times$ & 4.51$\times$ \\
Powerplant & 4.55M & 0.68 & 32.3 & 10.5M & 1.67 & 6.2 & 0.43$\times$ & 5.18$\times$ \\
\midrule
\textbf{Average} & \textbf{3.07M} & \textbf{2.41} & \textbf{108.9} & \textbf{10.7M} & \textbf{4.14} & \textbf{51.8} & \textbf{0.29$\times$} & \textbf{2.71$\times$} \\
\bottomrule
\end{tabular}
\caption{Computational cost comparison. Top 7 datasets are from the main evaluation; bottom 3 are additional benchmarks. \emph{Params}: model parameters. \emph{Train}: total end-to-end training time in hours. \emph{Sample}: time to generate one synthetic dataset in seconds. GATD uses 3.5$\times$ fewer parameters on average and requires 2.7$\times$ longer single-dataset sampling time due to 1000 steps (vs.\ TabDiff's 50). Total training runtime is 1.7$\times$ faster on average despite 2.5$\times$ more epochs. All timings are wall-clock measurements on a single NVIDIA T4 GPU.}
\label{tab:compute}
\end{table*}

\FloatBarrier

\subsection{Hyperparameters}
\label{app:hyperparams}

Table~\ref{tab:hyperparams} lists all hyperparameters used in our experiments. The configuration is identical across all 10 datasets and all three diffusion backbones, with the single exception of the architecture-depth choice: $n_{\text{blocks}} = 0$ for classification tasks and $n_{\text{blocks}} = 8$ for regression tasks (motivated in Section~\ref{sec:analysis} and analyzed empirically in Appendix~\ref{app:architecture}). Classification additionally uses a higher learning rate ($5 \times 10^{-3}$ vs.\ $1 \times 10^{-3}$) given the smaller backbone; the linear-decay schedule, batch size of 4{,}096, and AdamW optimizer with EMA (decay 0.9995, warmup 200 steps) are shared. The geometric loss weights $(\lambda_\theta, \lambda_\ell, \lambda_c) = (15, 15, 8)$ are likewise dataset-agnostic; per-dataset tuning unlocks additional gains on specific datasets (Appendix~\ref{app:lambda_sensitivity}, Section~\ref{sec:practitioner_tuning}) but is not required for the headline cross-architecture results.

\begin{table*}[h]
\centering
\small
\setlength{\tabcolsep}{5pt}
\begin{tabular}{lcc|lcc}
\toprule
\multicolumn{3}{c|}{\textbf{Architecture / Training}} &
\multicolumn{3}{c}{\textbf{EMA / Loss / Diffusion}} \\
\cmidrule(r){1-3} \cmidrule(l){4-6}
Hyperparameter & Classification & Regression &
Hyperparameter & Classification & Regression \\
\midrule

$d_{\text{model}}$ & 256 & 256 &
Decay & 0.9995 & 0.9995 \\

$n_{\text{blocks}}$ & 0 & 8 &
Warmup steps & 200 & 200 \\

Expansion factor & 4 & 4 &
$\lambda_\epsilon$ & 1.0 & 1.0 \\

Dropout & 0.1 & 0.1 &
$\lambda_{\text{cat}}$ & 0.05 & 0.05 \\

Time embedding dim & 128 & 128 &
$\lambda_\theta$ & 15.0 & 15.0 \\

Learning rate & $5 \times 10^{-3}$ & $1 \times 10^{-3}$ &
$\lambda_\ell$ & 15.0 & 15.0 \\

LR schedule & Linear decay & Linear decay &
$\lambda_c$ & 8.0 & 8.0 \\

Final LR & $10^{-6}$ & $10^{-6}$ &
$\sigma_{\min}$ & 0.002 & 0.002 \\

Batch size & 4096 & 4096 &
$\sigma_{\max}$ & 80 & 80 \\

Epochs & 20{,}000 & 20{,}000 &
$\sigma_{\text{data}}$ & 1.0 & 1.0 \\

Best model epoch & $\geq$10{,}000 & $\geq$10{,}000 &
Sampling steps & 1000 & 1000 \\

Gradient clipping & 1.0 & 1.0 &
& & \\

Weight decay & 0 & 0 &
& & \\

\bottomrule
\end{tabular}
\caption{Full hyperparameter settings.}
\label{tab:hyperparams}
\end{table*}

\subsection{Sampling-Step Ablation}
\label{app:steps_ablation}

We evaluated MLP+Geom against TabDiff at sampling-step budgets of 50, 100, 250, 500, 1000, and 2000 steps to characterize the sensitivity of GATD's reported gains to the sampling protocol. Per-(dataset, step-count) cells use \emph{single training seed and 20 generation seeds}, with the exception of the 1000-step row, which is reported from the main-results $3 \times 20$ protocol. The full $3 \times 20$ protocol was prohibitive across all six step-count cells. We compare each GATD step budget against TabDiff at its published 50-step protocol.

\begin{table*}[!htpb]
\centering
\small
\caption{GATD-MLP at varying sampling-step budgets vs TabDiff at its published 50-step protocol. Most rows use a single training seed and 20 generation seeds per cell ($1 \times 20$ protocol); the full $3 \times 20$ protocol was prohibitive across this many step-count cells. The 1000-step row, marked with $^*$, uses values from the main-results protocol ($3 \times 20$, 60 obs per cell). Aggregate values are means across all 10 datasets; MLE-1 and MLE-2 are reported as average percentage gap to real-data performance ($\downarrow$ better). Win counts (Shape/Trend/MLE-1/MLE-2 out of 10) compare GATD at each step budget against TabDiff at 50 steps. Performance improves monotonically up to 2000 steps with no clear saturation point in our evaluated range.}
\label{tab:step_ablation}
\begin{tabular}{lccccc}
\toprule
Configuration & Shape $\downarrow$ & Trend $\downarrow$ & MLE-1 gap $\downarrow$ & MLE-2 gap $\downarrow$ & Wins (S/T/M1/M2) \\
\midrule
TabDiff (50 steps) & 1.187 & 2.048 & 11.4\% & 9.7\% & --- \\
\midrule
GATD (50 steps) & 1.546 & 1.957 & 4.6\% & 7.2\% & 3/4/5/8 \\
GATD (100 steps) & 1.125 & 1.714 & 4.9\% & 5.4\% & 7/6/8/9 \\
GATD (250 steps) & 0.920 & 1.611 & 5.3\% & 6.4\% & 8/7/5/8 \\
GATD (500 steps) & 0.880 & 1.647 & 6.0\% & 7.0\% & 8/6/5/8 \\
GATD (1000 steps)$^*$ & 0.862 & 1.647 & 4.8\% & 5.6\% & 8/7/6/9 \\
GATD (2000 steps) & 0.845 & 1.577 & 4.1\% & 5.5\% & 8/7/7/9 \\
\bottomrule
\end{tabular}
\end{table*}

\begin{figure*}[!tpbh]
\centering
\includegraphics[width=\textwidth]{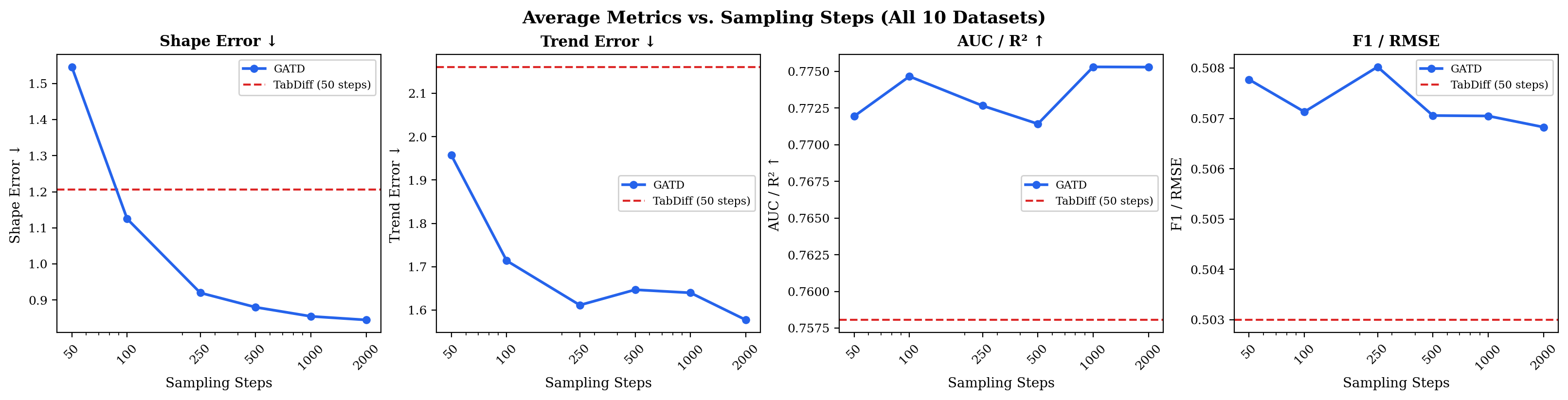}
\caption{\textbf{Sampling-step ablation summary.} Aggregate metrics across all 10 datasets as a function of GATD-MLP sampling steps, against TabDiff at its published 50-step protocol (red dashed lines). From left: Shape error ($\downarrow$), Trend error ($\downarrow$), MLE-1 gap to real ($\downarrow$), MLE-2 gap to real ($\downarrow$). GATD wins 3 of 4 aggregate metrics at 50 steps and all 4 by 100 steps; performance continues to improve up to 2000 steps with no clear saturation in our evaluated range. Single training seed and 20 generation seeds per (dataset, step-count) cell, except the 1000-step row which uses the main-results $3 \times 20$ protocol. Wall-clock costs in Section~\ref{app:steps_ablation}, paragraph ``Wall-clock cost.''}
\label{fig:step_ablation_summary}
\end{figure*}

\paragraph{Aggregate findings.} Table~\ref{tab:step_ablation} (visualized in Figure~\ref{fig:step_ablation_summary}) reports per-step-budget performance. At 50 sampling steps, GATD already wins 3 of 4 metrics in aggregate against TabDiff (Trend 1.96 vs 2.05; MLE-1 4.6\% vs 11.4\% gap; MLE-2 7.2\% vs 9.7\% gap), losing only on Shape (1.55 vs 1.19). By 100 sampling steps, GATD wins all 4 aggregate metrics. The story tightens further as step count increases: aggregate Shape error decreases monotonically from 1.55 (50 steps) to 0.85 (2000 steps), and aggregate Trend error decreases monotonically from 1.96 to 1.58. The MLE downstream-utility gaps remain in a narrow band (4--7\%) across all step counts, indicating that step count primarily affects distributional fidelity rather than predictive utility.

\paragraph{Per-dataset wins are mixed at low step counts.} While the aggregate at 50 steps shows GATD winning 3 of 4 metrics, the per-dataset story is mixed: Shape 3/10, Trend 4/10, MLE-1 5/10, MLE-2 8/10. By 100 steps, per-dataset wins improve to 7/6/8/9; by 250 steps and beyond, the pattern stabilizes around 8/7/5--7/8--9. The aggregate-vs-per-dataset divergence at low step counts reflects a few large-effect datasets pulling aggregate means below TabDiff's, while the median dataset still favors TabDiff at very low step budgets.

\paragraph{No saturation point in our evaluated range.} Our data does not show a clear saturation point: aggregate Shape continues to improve from 0.86 at 1000 steps to 0.85 at 2000 steps, and aggregate Trend from 1.65 to 1.58. The 1000-step setting we report as default in the main text is a reasonable operating point for accuracy-vs-cost trade-off, not a saturation point---practitioners willing to spend additional inference cost can extract modest additional gains. Conversely, at 50--100 steps, GATD is already beating TabDiff on most aggregate metrics, suggesting that practitioners constrained by inference budget can also use GATD without giving up most of its advantage.

\paragraph{Wall-clock cost.} A representative example from the Adult dataset: TabDiff requires 35 seconds to generate a single synthetic dataset at 50 steps, whereas GATD-MLP requires 2.5 seconds at 50 steps and 5 seconds at 100 steps---an order of magnitude faster per-step due to the substantially smaller MLP backbone. At 1000 steps GATD-MLP requires roughly 50 seconds per dataset, comparable to TabDiff at 50 steps; at 2000 steps the cost is roughly 100 seconds.

\FloatBarrier

\subsection{Wide-Table Scalability: APS Failure at Scania Trucks}
\label{app:aps_scalability}

To empirically characterize GATD's behavior on wider tables than those in the main 10-dataset benchmark, we evaluated against TabDiff on the APS Failure at Scania Trucks dataset ($d = 171$ columns, $n = 60{,}000$ rows, 1 categorical column). This dataset sits well above the $d \leq 48$ regime of our main evaluation and tests the practical implications of $O(d^2)$ pairwise feature scaling. Both methods use default configurations and train on a single NVIDIA T4 GPU.

\paragraph{Default configuration.} With the same loss weights $(\lambda_\theta, \lambda_\ell, \lambda_c) = (15, 15, 8)$ used across the main 10-dataset benchmark, GATD already substantially outperforms TabDiff on distributional fidelity at $d = 171$ (Table~\ref{tab:aps_default}): Shape error drops from 9.48\% to 0.51\% (95\% relative reduction), and Trend is essentially tied. TabDiff retains the AUC advantage on this single-categorical-column dataset; the small geometric signal available from a single discrete column appears insufficient to recover downstream classification utility under the default loss weighting, although tuned weights (below) substantially close this gap. The $O(d^2)$ pairwise feature cost is empirically observable: at $d = 171$, GATD's parameter count (17M vs.\ 12M) and memory footprint (12.3 GB vs.\ 6.5 GB) exceed TabDiff's, but both models fit on a single T4 and GATD is still slightly faster per epoch (5.5s vs.\ 7.5s).
\begin{table*}[h]
\centering
\small
\begin{tabular}{lcc}
\toprule
& GATD (default 15/15/8) & TabDiff \\
\midrule
Shape Error $\downarrow$ & \textbf{0.51\%} & 9.48\% \\
Trend Error $\downarrow$ & \textbf{9.15\%} & 9.56\% \\
AUC $\uparrow$ & 87.0\% & \textbf{97.9\%} \\
\midrule
Training (s/epoch) & \textbf{5.5} & 7.5 \\
Memory (GB) & 12.3 & \textbf{6.5} \\
Parameters & 17M & \textbf{12M} \\
\bottomrule
\end{tabular}
\caption{APS Failure at Scania Trucks ($d = 171$), default-hyperparameter comparison. With default $(\lambda_\theta, \lambda_\ell, \lambda_c) = (15, 15, 8)$, GATD substantially outperforms TabDiff on distributional fidelity (Shape 95\% relative reduction; Trend essentially tied), while TabDiff wins downstream utility (AUC), consistent with APS having only 1 categorical column among 171. The $O(d^2)$ scaling cost is empirically observable at this dimensionality: GATD's parameter count and memory usage exceed TabDiff's, though both fit on a single T4 and GATD remains slightly faster per epoch.}
\label{tab:aps_default}
\end{table*}

\paragraph{With minimal tuning.} Following the practitioner guidance in Section~\ref{sec:practitioner_tuning}, we performed a small grid search over $(\lambda_\theta, \lambda_\ell, \lambda_c) \in \{0, 60\}^3$ (8 trials total), motivated by the observation that APS has only 1 categorical column and should benefit from emphasizing length supervision over angle supervision. The grid identifies $(0, 60, 0)$ as the optimal configuration. Table~\ref{tab:aps_tuned} reports the tuned-vs-TabDiff comparison: GATD now wins 3 of 4 metrics, including a substantial F1 margin (73.2\% vs.\ 44.8\%) where the default configuration had lost AUC. The tuning cost (8 trials) is modest, and demonstrates that the practitioner-tunable $\lambda$ values address failure modes that TabDiff's fixed-architecture protocol cannot reach.

\begin{table*}[h]
\centering
\small
\begin{tabular}{lcc}
\toprule
& GATD (tuned: 0/60/0) & TabDiff \\
\midrule
Shape Error $\downarrow$ & \textbf{0.64\%} & 9.48\% \\
Trend Error $\downarrow$ & \textbf{9.22\%} & 9.56\% \\
AUC $\uparrow$ & 95.1\% & \textbf{97.9\%} \\
F1 $\uparrow$ & \textbf{73.2\%} & 44.8\% \\
\bottomrule
\end{tabular}
\caption{APS with 8-trial loss-weight grid search. GATD wins 3 of 4 metrics, including a substantial F1 margin (73.2\% vs 44.8\%). The tuning cost (8 trials) is modest relative to the gains, and demonstrates that the practitioner-tunable lambdas address failure modes that TabDiff's fixed-architecture protocol cannot.}
\label{tab:aps_tuned}
\end{table*}

\paragraph{Interpretation.} The $O(d^2)$ pairwise feature scaling is real and observable at this dimensionality---GATD's memory consumption and parameter count both exceed TabDiff's at $d = 171$. However, the practical implications are not the catastrophic failure suggested by an asymptotic complexity argument: training time per epoch remains favorable, both methods fit comfortably on a single T4, and tuned GATD wins 3 of 4 metrics. This is also the same order as column-wise self-attention, so the relevant question is not whether quadratic pair structure exists, but whether the pairwise signal is useful. Our results and failed pair-selection experiments support supervising all pairs, including weak or noisy relationships, because those non-relationships are part of the relational signature.

\FloatBarrier

\subsection{GATD at TabDiff's 8{,}000-Epoch Training Protocol}
\label{app:gatd_8k_epochs}

Reviewers raised a reasonable concern that GATD's 20{,}000-epoch training schedule (vs.\ TabDiff's published 8{,}000) might be a confounding source of GATD's reported gains rather than the geometry itself. To isolate geometric supervision from training-protocol differences, we additionally trained GATD-MLP at 8{,}000 epochs---identical to TabDiff's protocol---across 3 training seeds and 20 generation seeds per dataset, holding other hyperparameters at defaults. Table~\ref{tab:gatd_8k_epochs} reports the result.

\paragraph{Interpretation.} Extended training contributes at most a small fraction of GATD's reported improvement over TabDiff. At identical 8{,}000-epoch budgets, GATD-MLP still achieves 27 of 40 wins (with 3 ties) against TabDiff, compared to 29 wins and 2 ties at 20{,}000 epochs. The 20{,}000-epoch protocol we report by default in the main text is a moderately favorable operating point for GATD but is not the source of its gains over TabDiff: the gap between GATD-8k (0.854 Shape) and TabDiff-8k (1.187 Shape) is 28\% relative reduction, very close to the 27\% reduction reported at 20k. The marginal benefit of 2.5$\times$ more training is small (Shape 0.854 $\to$ 0.862, a 0.9\% change; MLE-2 gap 6.1\% $\to$ 5.6\%). This isolates geometric supervision (and architectural simplification to an MLP) as the primary mechanisms, not the extended training budget.

\begin{table*}[h]
\centering
\small
\begin{tabular}{lcccc}
\toprule
Configuration & Shape $\downarrow$ & Trend $\downarrow$ & MLE-1 gap $\downarrow$ & MLE-2 gap $\downarrow$ \\
\midrule
TabDiff (8k epochs, published) & 1.187 & 2.048 & 11.4\% & 9.7\% \\
GATD (8k epochs, matched) & \textbf{0.854} & \textbf{1.68} & \textbf{4.7\%} & \textbf{6.1\%} \\
GATD (20k epochs, default) & 0.862 & 1.65 & 4.8\% & 5.6\% \\
\bottomrule
\end{tabular}
\caption{GATD-MLP at TabDiff's 8{,}000-epoch training budget. At identical training budgets, GATD wins 27 of 40 metric-dataset cells with 3 ties, compared to 29 wins and 2 ties at 20{,}000 epochs. The marginal benefit of 2.5$\times$ more training is small: Shape error changes only from 0.854 to 0.862, while MLE-2 gap decreases from 6.1\% to 5.6\%. Geometric supervision, not extended training, is the primary driver of GATD's improvement over TabDiff.}
\label{tab:gatd_8k_epochs}
\end{table*}

\FloatBarrier

\subsection{Loss-Weight Sensitivity Analysis}
\label{app:lambda_sensitivity}

The default geometric loss weights $(\lambda_\theta, \lambda_\ell, \lambda_c) = (15, 15, 8)$ are not finely tuned: they were chosen at the outset of development based on rough alignment between angle/length and diffusion loss magnitudes, and held fixed across all 10 datasets and all three diffusion backbones. To characterize whether this default sits in a stable basin or at a precarious peak, we conducted single-variable sensitivity sweeps, holding the other two weights at default values:

\begin{itemize}[leftmargin=*,nosep]
\item \textbf{Angle sweep (primary):} $\lambda_\theta \in \{0, 5, 10, 15, 20, 30\}$ with $\lambda_\ell = 15$, $\lambda_c = 8$ fixed.
\item \textbf{Length sweep:} $\lambda_\ell \in \{0, 5, 10, 15, 20, 30\}$ with $\lambda_\theta = 15$, $\lambda_c = 8$ fixed.
\item \textbf{Consistency sweep:} $\lambda_c \in \{0, 5, 10, 15, 20, 30\}$ with $\lambda_\theta = 15$, $\lambda_\ell = 15$ fixed.
\end{itemize}

Sweeps were run on 4 representative datasets, each cell at a \emph{single training seed and a single generation seed} ($1 \times 1$ protocol). The full $3 \times 20$ protocol used for the main results was prohibitive at the scale of this sensitivity grid: $6 \text{ values} \times 3 \text{ lambdas} \times 3 \text{ train seeds} \times 20 \text{ gen seeds} \times 10 \text{ datasets} = 10{,}800$ generations, well outside the available compute budget.

\begin{figure*}[!htpb]
\centering
\includegraphics[width=\textwidth]{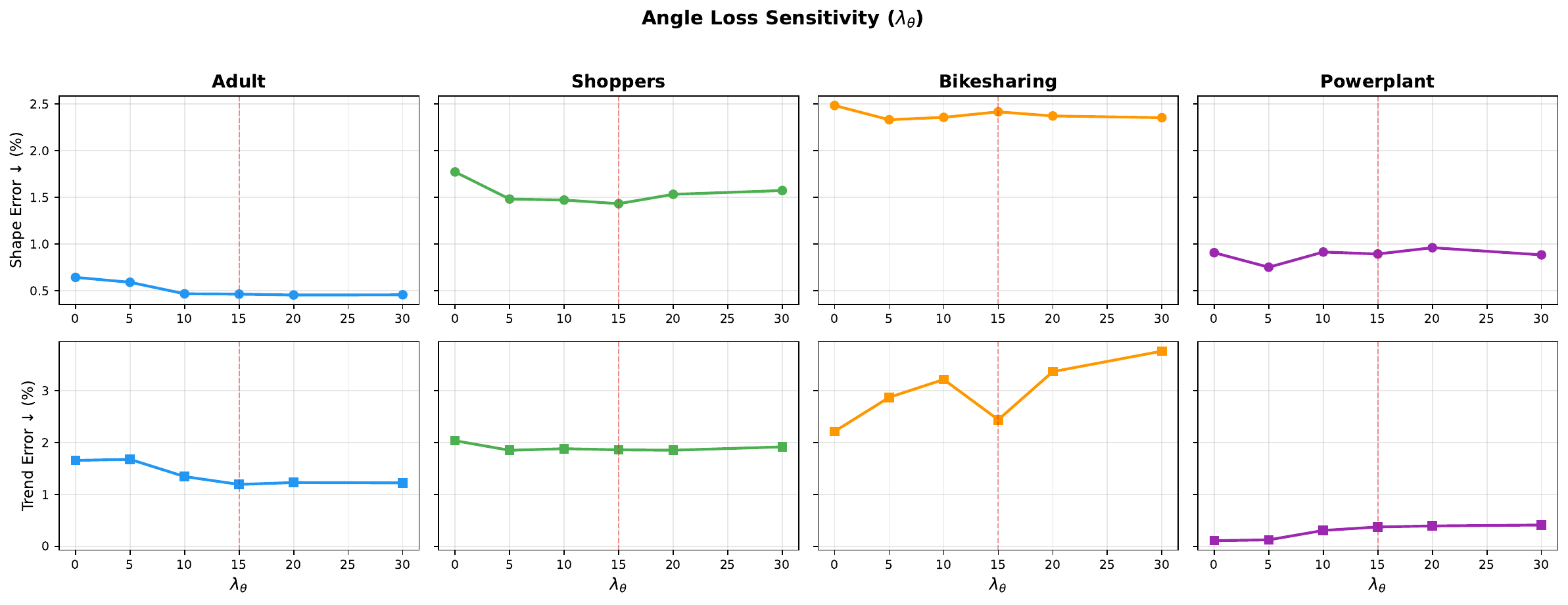}
\includegraphics[width=\textwidth]{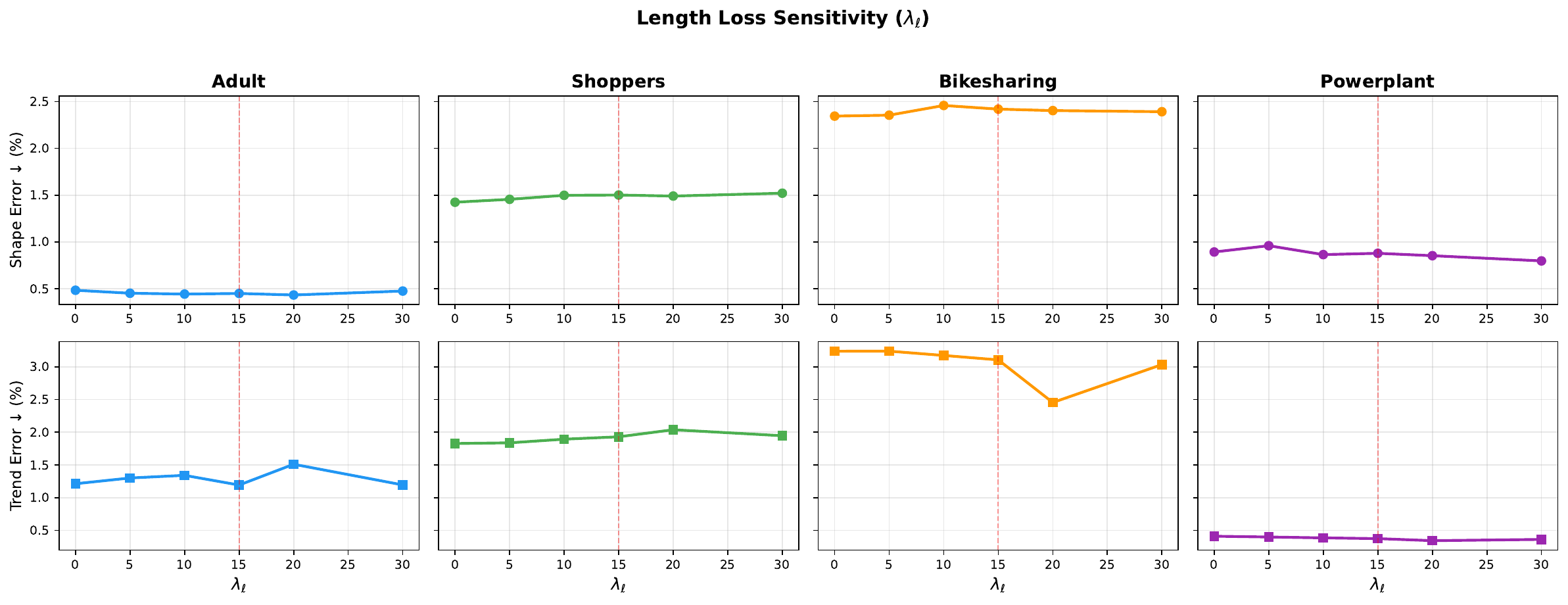}
\includegraphics[width=\textwidth]{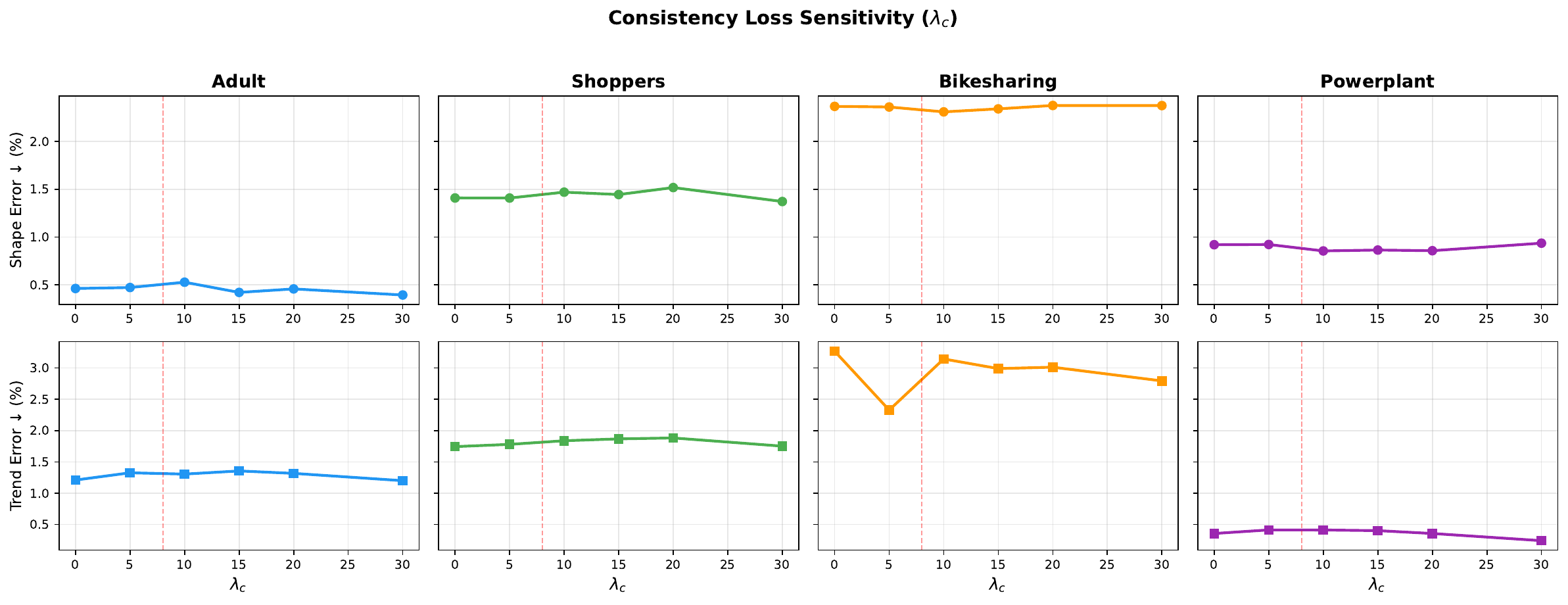}
\caption{Loss-weight sensitivity sweeps for the three geometric loss terms on four representative datasets. Each panel varies one weight while holding the other two at the default setting: angle sweep $(\lambda_\theta, \lambda_\ell, \lambda_c)=(X,15,8)$, length sweep $(15,X,8)$, and consistency sweep $(15,15,X)$, with $X \in \{0,5,10,15,20,30\}$ under the $1\times1$ protocol. Aggregate Shape and Trend errors remain within $\pm5\%$ of the default $(15,15,8)$ across the sweeps. Per-dataset behavior is heterogeneous: continuous-heavy datasets such as Powerplant can improve when angle supervision is reduced, whereas categorical-heavy datasets such as Adult generally prefer the default or higher angle weight.}
\label{fig:lambda_sweep}
\end{figure*}

Across angle, length, and consistency sweeps on these 4 datasets (Figure~\ref{fig:lambda_sweep}), single-seed Shape and Trend errors remain within close range of the default $(15, 15, 8)$ configuration over the full $\lambda_\theta \in \{0, 30\}$ range. The default sits in a stable basin rather than at a precarious peak, indicating that GATD does not require precise hyperparameter calibration to achieve its reported performance.

\paragraph{Per-dataset tuning unlocks additional gains.} Although the default is stable in aggregate, per-dataset tuning of the geometric weights unlocks substantial additional improvements. The most striking example is Powerplant (0 categorical columns, $d = 5$): at default $(15, 15, 8)$, Powerplant achieves 0.973 Shape error and 0.387 Trend error. Setting $\lambda_\theta = 0$---disabling the angle loss---yields 0.908 Shape and 0.108 Trend, halving TabDiff's best Trend result on this dataset (0.219). A small grid search over $\lambda_\theta, \lambda_\ell \in \{0, 30, 60\}$ (9 configurations) identifies $(\lambda_\theta, \lambda_\ell, \lambda_c) = (0, 60, 60)$ as the optimal: 0.764 Shape, 0.111 Trend on Powerplant. These per-dataset tunable knobs have no equivalent in TabDiff or other transformer-based tabular generators, where inter-column relationships are learned implicitly without explicit user-tunable weights.

The pattern parallels the per-dataset improvements observed on the MLP backbone (Section~\ref{sec:analysis}, Appendix~\ref{app:cat_analysis}): on low-categorical datasets, disabling the angle loss and emphasizing length supervision can improve aggregate metrics. We provide concrete tuning recommendations in Section~\ref{sec:practitioner_tuning}.


\subsection{Practitioner Guidance for Loss-Weight Tuning}
\label{sec:practitioner_tuning}

The geometric loss weights $(\lambda_\theta, \lambda_\ell, \lambda_c)$ provide tunable practitioner levers absent from transformer-based tabular generators. Our default configuration $(15, 15, 8)$ achieves competitive performance across all 10 datasets without per-dataset tuning, but per-dataset tuning unlocks substantial additional gains in some regimes. Based on our sensitivity analysis (Appendix~\ref{app:lambda_sensitivity}) and the categorical-anchor analysis (Section~\ref{sec:analysis}), we offer the following practitioner guidance:

\begin{itemize}[leftmargin=*,nosep]
\item \textbf{Start with the default $(15, 15, 8)$.} The default sits in a stable performance basin: aggregate Shape and Trend errors remain within $\pm 5\%$ of the default across the full $\lambda \in \{0, 30\}$ range when sweeping any single weight (Appendix~\ref{app:lambda_sensitivity}).
\item \textbf{Categorical-heavy datasets (high categorical-column fraction):} the angle loss is empirically the primary driver on the MLP backbone (Section~\ref{sec:analysis}). Increasing $\lambda_\theta$ to 20--30 may improve Shape on datasets with rich categorical structure.
\item \textbf{Continuous-heavy datasets (low categorical-column fraction):} the length loss is the primary driver. Increasing $\lambda_\ell$ and reducing $\lambda_\theta$ (potentially to 0) may unlock substantial gains. On Powerplant (0 categorical columns), reducing $\lambda_\theta$ from 15 to 0 cuts Trend error by 72\% (0.387 to 0.108), beating TabDiff's best Trend on this dataset (0.219) by a factor of two.
\item \textbf{Wide tables with few categorical columns:} prefer $(0, 60, 0)$ as a starting point. On APS Failure at Scania Trucks ($d = 171$, 1 categorical column; Appendix~\ref{app:aps_scalability}), this configuration achieves 3 of 4 wins against TabDiff after only an 8-trial grid search.
\end{itemize}

These knobs are not available in TabDiff or other transformer-based tabular generators, where inter-column relationships are learned implicitly without explicit user-tunable weights. The geometric loss weights thus provide an additional dimension of practitioner control that complements the architecture-portability claim of Section~\ref{sec:main_results}.


\FloatBarrier

\subsection{Geometric Feature Form: Arctan vs Raw Differences}
\label{app:arctan_vs_diffs}

To validate the arctan transformation as the geometric feature choice, we compared GATD-MLP against an otherwise-identical configuration using raw pairwise differences ($v_j - v_i$) as the angle feature, with the length feature unchanged. All other architecture, training, and supervision details are held identical. Per-dataset cells use a \emph{single training seed and 20 generation seeds} ($1 \times 20$ protocol); the main-text $3 \times 20$ protocol was not run for this ablation.

\begin{table*}[h]
\centering
\small
\caption{Arctan vs raw pairwise differences as the angle feature, evaluated on all 10 datasets. Arctan = the default GATD configuration. RawDiffs = identical architecture, training, and supervision with $\theta_{ij} = \arctan(v_j - v_i)$ replaced by $\theta_{ij} = v_j - v_i$. Single training seed, 20 generation seeds per cell ($1 \times 20$ protocol). Both feature forms produce competitive results; arctan provides a consistent edge, particularly on downstream-utility metrics.}
\label{tab:arctan_vs_diffs}
\begin{tabular}{lccc}
\toprule
Metric & Arctan (avg) & RawDiffs (avg) & Wins (Arctan / RawDiffs) \\
\midrule
Shape Error $\downarrow$ & 0.862 & 0.906 & 7/3 \\
Trend Error $\downarrow$ & 1.647 & 1.664 & 5/5 \\
MLE-1 (gap to real) $\downarrow$ & 4.77\% & 5.07\% & 8/2 \\
MLE-2 (gap to real) $\downarrow$ & 5.62\% & 6.08\% & 8/2 \\
\bottomrule
\end{tabular}
\end{table*}
\FloatBarrier

\paragraph{Findings.} Both feature forms produce competitive results (Table~\ref{tab:arctan_vs_diffs}), confirming that pairwise auxiliary supervision is the primary mechanism rather than the specific transformation choice. Arctan, however, provides a consistent edge on Shape (7/10 datasets, aggregate 0.86 vs 0.91), MLE-1 (8/10, aggregate 4.8\% vs 5.1\% gap), and MLE-2 (8/10, aggregate 5.6\% vs 6.1\% gap). On Trend the two forms are essentially tied at the per-dataset level (5/5), with arctan winning slightly in aggregate (1.65 vs 1.66).

\paragraph{Why arctan helps.} Two properties of arctan provide consistent benefits over raw differences. First, arctan's bounded output ($\theta_{ij} \in (-\pi/2, \pi/2)$) provides stable supervision targets regardless of input scale, whereas raw differences inherit the scale of the underlying column values. Second, arctan's nonlinear compression (large differences map to similar angles near $\pm\pi/2$) emphasizes the directional structure of column relationships over their absolute magnitudes, which is precisely the inductive bias the geometric supervision is designed to provide. These properties matter most for distributional fidelity (Shape) and downstream utility (MLE-1, MLE-2) where the bounded representation appears to provide more consistent gradients during training.

The arctan formulation also opens a natural extension to higher-order geometric features such as triangle-closure losses on column triples (Section~\ref{sec:conclusion}), which raw differences would not support without additional normalization.

\end{document}